\documentclass[10pt,twocolumn,twoside]{IEEEtran}
\ifCLASSINFOpdf \else \fi

\usepackage{graphicx}
\usepackage{verbatim}
\usepackage[cmex10]{amsmath}
\usepackage{amssymb}
\usepackage{amsfonts}
\usepackage{bbm}

\usepackage{color}

\usepackage{algorithmic}
\usepackage{array}
\usepackage{multirow}
\ifCLASSOPTIONcompsoc
 \usepackage[caption=false,font=normalsize,labelfont=sf,textfont=sf]{subfig}
\else
 \usepackage[caption=false,font=footnotesize]{subfig}
\fi
\usepackage{url}
\usepackage[noadjust]{cite}

\begin{document}

\title{AgentGraph: Towards Universal Dialogue Management with Structured Deep Reinforcement Learning}

\author{{Lu Chen,~\IEEEmembership{Student Member,~IEEE}, Zhi Chen, Bowen Tan, Sishan Long,\\ Milica Ga{\v{s}}i{\'c},~\IEEEmembership{Member,~IEEE}, and Kai Yu,~\IEEEmembership{Senior Member,~IEEE}}
\thanks{Lu Chen, Zhi Chen, Bowen Tan, Sishan Long, and Kai Yu are supported by the National Key Research and Development Program of China under Grant No.2017YFB1002102, and Shanghai International Science and Technology Cooperation Fund (No. 16550720300). Milica Ga{\v{s}}i{\'c} is supported by an Alexander von Humboldt Sofja Kovalevskaja award. 
 (Corresponding author: Kai Yu)

Lu Chen, Zhi Chen, Bowen Tan, Sishan Long, and Kai Yu are with the Department of  Computer Science and Engineering, Shanghai Jiao Tong University, Shanghai 200240, China (e-mail:chenlusz@sjtu.edu.cn; zhenchi713@sjtu.edu.cn; tanbowen@sjtu.edu.cn; longsishan@sjtu.edu.cn; kai.yu@sjtu.edu.cn).

Milica Ga{\v{s}}i{\'c} is with Heinrich Heine University D\"{u}sseldorf, Germany (e-mail: gasic@uni-duesseldorf.de).
}}


\maketitle

\begin{abstract}

Dialogue policy plays an important role in task-oriented spoken dialogue systems. It determines how to respond to users. The recently proposed deep reinforcement learning (DRL) approaches have been used for policy optimization. However, these deep models are still challenging for two reasons: 1) Many DRL-based policies are not sample-efficient. 2) Most models don't have the capability of policy transfer between different domains.
In this paper, we propose a universal framework, \textit{AgentGraph}, to tackle these two problems. The proposed AgentGraph is the combination of GNN-based architecture and DRL-based algorithm. It can be regarded as one of the multi-agent reinforcement learning approaches. Each agent corresponds to a node in a graph, which is defined according to the dialogue domain ontology. When making a decision, each agent can communicate with its neighbors on the graph. 
Under AgentGraph framework, we further propose Dual GNN-based dialogue policy, which implicitly decomposes the decision in each turn into a high-level global decision and a low-level local decision.
Experiments show that AgentGraph models significantly outperform traditional reinforcement learning approaches on most of the 18 tasks of the PyDial benchmark. Moreover, when transferred from the source task to a target task, these models not only have acceptable initial performance but also converge much faster on the target task.
\end{abstract}

\begin{IEEEkeywords}
dialogue policy, deep reinforcement learning, graph neural networks, policy adaptation, transfer learning
\end{IEEEkeywords}

\IEEEpeerreviewmaketitle

\section{Introduction}

Nowadays, conversational systems are increasingly used in smart devices, e.g.  Amazon Alexa, Apple Siri, and Baidu Duer. One feature of these systems is that they can  interact with humans through speech to accomplish a task, e.g. booking a movie ticket or finding a hotel. This kind of systems are also called task-oriented spoken dialogue systems (SDS).
They usually consist of three components: input module, control module, and output module. The control module is also referred to as {\em dialogue management} (DM) \cite{young2010hidden}. It is the core of whole system. It has two missions: one is to maintain the dialogue state, and another is to decide how to respond according to a dialogue policy, which is the focus of this paper.

In commercial dialogue systems, the dialogue policy is usually defined as some hand-crafted rules in the form of mapping dialogue states to actions. This is known as rule-based dialogue policy. However, in real-life applications,  noises from the input module\footnote{The input module usually includes automatic speech recognition (ASR) and spoken language understanding (SLU).} are inevitable, which makes true dialogue state is unobservable. It is questionable as to whether the rule-based policy can handle this kind of uncertainty. Hence, statistical dialogue management is proposed and attracts lots of research interests in the past few years. The partially observable Markov decision process (POMDP) provides a well-founded framework for statistical DM \cite{young2010hidden,Young:2013gt}.

Under the POMDP-based framework, at every dialogue turn, belief state $\mathbf{b}_t$, i.e. a distribution of possible states, is updated according to last belief state and current input. Then reinforcement learning (RL) methods automatically optimize the policy $\pi$, which is a function from belief state $\mathbf{b}_t$ to dialogue action $a_t=\pi(\mathbf{b}_t)$~\cite{Young:2013gt}. Initially, linear RL-based models are adopted, e.g. natural actor-critic \cite{thomson2010bayesian,jurvcivcek2011natural}. However, these linear models have a poor ability of expression and suffer from slow training. Recently, nonparametric algorithms, e.g. Gaussian process reinforcement learning (GPRL) \cite{Gasic:2010tl,gasic-tasl2014}, have been proposed.  They can be used to optimize policies from a small number of dialogues.  However, the computation cost of these nonparametric models increases with the increase of the number of data. 
As a result, these methods cannot be used in large-scale commercial dialogue systems \cite{su2017sample}.

More recently, deep neural networks are utilized for the approximation of dialogue policy, e.g. deep $Q$-networks and policy networks \cite{cuayahuitl-nipsworkship2015,Fatemi:2016tr,zhao-sigdial2016,williams2017hybrid,chen-emnlp2017,chang2017affordable,li2017end,Weisz2018}. These models are known as deep reinforcement learning (DRL), which is often more expressive and computationally effective than traditional RL.  However, these deep models are still challenging for two reasons. 
\begin{itemize}
    \item First, traditional DRL-based methods are not sample-efficient, i.e. thousands of dialogues are needed for training an acceptable policy. Therefore on-line training dialogue policy with real human users is very costly. 
    \item Second, unlike GPRL \cite{gavsic2015policy,gavsic2013pomdp}, most DRL-based policies cannot be transferred between different domains. The reason is that the ontologies of the two domains usually are fundamentally different, resulting in different dialogue state spaces and action sets, which means the input space and output space of two DRL-based policies have to be different.
\end{itemize}

In this paper, we propose a {\it universal} framework with {\it structured} deep reinforcement learning to address the above problems.
The framework is based on graph neural networks (GNN) \cite{scarselli2009graph} and  is called {\bf AgentGraph}. It consists of some sub-networks, each one corresponding to a node of a directed graph, which is defined according to the domain ontology including slots and their relations.
The graph has two types of nodes: {\em slot-independent} node (I-node) and {\em slot-dependent} node (S-node). Each node can be considered as a sub-agent\footnote{In this paper, we use node/S-node/I-node and agent/S-agent/I-agent interchangeably.}. The same types of nodes share parameters. This can improve the speed of policy learning. In order to model the interaction between agents, each agent can communicate with its neighbors when making a decision. 
Moreover, when a new domain appears, the shared parameters of S-agents and the parameters of I-agent in the original domain can be used to initialize the parameters of AgentGraph in the new domain. 

The initial version of AgentGraph is proposed in our previous work \cite{chen2017policy,chen2018structured}. Here we give a more comprehensive investigation of this framework from four-fold: 1) The Domain Independent Parametrization (DIP) function \cite{wang2015learning} is used to abstract belief state. The use of DIP avoids the use of private parameters for each agent, which is beneficial to the domain adaptation. 2) Besides the vanilla GNN-based policy, we propose a new architecture of AgentGraph, i.e. Dual GNN (DGNN)-based policy. 3) We investigate three typical graph structures and two message commutation methods between nodes in the graph. 4) Our proposed framework is evaluated in PyDial benchmark. It not only performs better than typical RL-based models on most tasks but also can be transferred across tasks.

The rest of the paper is organized as follows. We first introduce statistical dialogue management in the next section. Then, we describe the details of AgentGraph in Section \ref{sec:agentgraph}. In Section \ref{sec:sdp} we propose two instances of AgentGraph for dialogue policy. This is followed with a description of policy transfer learning under AgentGraph framework in Section \ref{sec:policy_transfer}. The results of the extensive experiments are given in Section \ref{sec:exp}. We conclude and give some future research directions in Section \ref{sec:conclusion}.

\section{Statistical Dialogue Management}
\label{sec:dm}
Statistical dialogue management can be cast as a partially observable Markov decision process (POMDP) \cite{Young:2013gt}. It is defined as a 8-tuple ($\mathbb{S}, \mathbb{A}, \mathbb{T}, \mathbb{O}, \mathbb{Z}, \mathbb{R},\gamma, \mathbf{b}_0$). $\mathbb{S}$ and $\mathbb{A}$ denote a set of dialogue states $s$ and a set of dialogue actions $a$ respectively. $\mathbb{T}$ defines transition probabilities between states $P(s_t|s_{t-1}, a_{t-1})$. 
$\mathbb{O}$ denotes a set of observations $o$.  $\mathbb{Z}$ defines an observation probability $P(o_t|s_t, a_{t-1})$. 
$\mathbb{R}$ defines the reward function $r(s_t, a_t)$. $\gamma$ is a discount factor with $\gamma \in (0,1]$, which decides how much immediate rewards are favored over future rewards.
$\mathbf{b}_0$ is an initial belief over possible dialogue states. 

At each dialogue turn, the environment is in some {\em unobserved} state $s_t$. The conversational agent receives an observation $o_t$ from the environment, and updates its belief dialogue state $\mathbf{b}_t$, i.e. a probability distribution over possible states. Based on $\mathbf{b}_t$, the agent selects a dialogue action $a_t$ according to a dialogue policy $\pi(\mathbf{b}_t)$, then obtains an immediate reward $r_t$ from the environment, and transitions to an unobserved state $s_{t+1}$.

\subsection{Belief Dialogue State Tracking}
In task-oriented conversational systems, the dialogue state is typically defined according to a structured {\em ontology} including some slots and their relations. Each slot can take a value from the candidate value set. The user intent can be defined as a set of slot-value pairs, e.g. \{{\em price=cheap, area=west}\}. It can be used as a constraint to frame a database query. 
Because of the noise from ASR and SLU modules, the agent doesn't exactly know the user intent. 
Therefore, at each turn, a dialogue state tracker maintains a probability distribution over candidate values for each slot, which is known as marginal belief state.
After the update of belief state, the values with the largest belief for each slot are used as a constraint to search the database. The matched entities in the database with other general features as well as the marginal belief states for slots are concatenated as whole belief dialogue state, which is the input of dialogue policy. Therefore, the belief state $\mathbf{b}_t$ usually can be factorized into some {\em slot-dependent} belief states and a {\em slot-independent} belief state, i.e. $\mathbf{b}_t=\mathbf{b}_{t0} \oplus \mathbf{b}_{t1}\oplus \cdots \oplus\mathbf{b}_{tn}$. $\mathbf{b}_{ti}(1\le\ i \le n)$\footnote{For simplicity, in following sections we will use $\mathbf{b}_{i}$ shorthand for $\mathbf{b}_{ti}$ when there is no confusion.} is the marginal belief state of $i$-th slot, and $\mathbf{b}_{t0}$ denotes the set of general features, which are usually slot-independent.
A various of models are proposed for dialogue state tracking (DST)  \cite{hendersonmachine,sun-slt14,yu-tasl15,ky219-yu-fcs15,sun-EtAl:2014:W14-43,xie-sigdial15}. 
The state-of-the-art methods utilize deep learning \cite{zhong2018global,ramadan2018large,mrkvsic2017neural,ren2018towards}.

\subsection{Dialogue Policy Optimization}
The dialogue policy $\pi(\mathbf{b}_t)$ decides how to respond to the users. The system actions $\mathbb{A}$ usually can be divided into $n+1$ sets, i.e. $\mathbb{A} = \mathbb{A}_0 \cup \mathbb{A}_1 \cup \cdots \cup \mathbb{A}_n$.  
$\mathbb{A}_{0}$ is the slot-independent action set, e.g. {\em inform}(), {\em bye}(), {\em restart}() \cite{young2010hidden}, and 
$\mathbb{A}_{i}(1\le i\le n)$ are slot-dependent action sets, e.g. {\em select}($slot_i$), {\em request}($slot_i$), {\em confirm}($slot_i$) \cite{young2010hidden}.

A conversational agent is trained to find an optimal policy that maximizes the expected discounted long-term return in each belief state $\mathbf{b}_t$:
\begin{equation}
V(\mathbf{b}_t) = \mathbb{E}_{\pi}(R_t|\mathbf{b}_t) = \mathbb{E}_{\pi}(\sum_{i=t}^T \gamma^{i-t} r_i).
\end{equation}
The quantity $V(\mathbf{b}_t)$ is also referred to as a {\em value function}. It tells how good the agent is to be in the belief state $\mathbf{b}_t$. A related quantity is the {\em Q-function} $Q(\mathbf{b}_t,a_t)$. It is the expected discounted long-term return by taking action $a_t$ in belief state $\mathbf{b}_t$, then following the current policy $\pi$: $Q(\mathbf{b}_t,a_t) = \mathbb{E}_{\pi}(R_t|\mathbf{b}_t,a_t)$. Intuitively, the $Q$-function measures how good a dialogue action $a_t$ is taken in the belief state $\mathbf{b}_t$. 
By definition, the relation between value function and $Q$-function is that $V(\mathbf{b}_t) = \mathbb{E}_{a_t\sim \pi(\mathbf{b}_t)} Q(\mathbf{b}_t,a_t)$. For a deterministic policy, the best action $a^* = \arg\max_{a_t} Q(\mathbf{b}_t,a_t)$, therefore 
\begin{equation}
V(\mathbf{b}_t) = \max_{a_t} Q(\mathbf{b}_t,a_t).
\label{eq:vq_relation}
\end{equation}
Another related quantity is the {\em advantage function}:
\begin{equation}
A(\mathbf{b}_t,a_t) = Q(\mathbf{b}_t,a_t) - V(\mathbf{b}_t).
\label{eq:advantage}
\end{equation}
It measures the relative importance of each action. Combining Equation (\ref{eq:vq_relation}) and Equation (\ref{eq:advantage}), we can obtain that $\max_{a_t} A(\mathbf{b}_t,a_t) = 0$ for a deterministic policy.

The state of the art statistical approaches for automatic policy optimization are based on RL \cite{Young:2013gt}. Typical RL-based methods include Gaussian process reinforcement learning (GPRL) \cite{Gasic:2010tl,gasic-tasl2014,chen-sigdial2015} and Kalman temporal difference~(KTD) reinforcement learning \cite{pietquin-ijcai-2011}. Recently, deep reinforcement learning (DRL) \cite{mnih-nature2015} has been investigated for dialogue policy optimization, 
e.g. Deep $Q$-Networks (DQN) \cite{cuayahuitl-nipsworkship2015,zhao-sigdial2016,lipton-arxiv2016,Fatemi:2016tr,chen-eacl2017,chang2017affordable,baolin-hier,chen2018structured}, policy gradient methods \cite{su2017sample,williams2017hybrid}, and actor-critic approaches \cite{Fatemi:2016tr}. 
However, compared with GPRL and KTD-RL, most of these deep models are not sample-efficient. More recently, some methods are proposed to improve the speed of policy learning based on improved DRL algorithms, e.g. eNAC \cite{su2017sample}, ACER~\cite{Weisz2018,Wang2016SampleEA} or  BBQN \cite{lipton-arxiv2016}. 
In contrast, here we take an {\em alternative} approach, i.e. we propose a {\em structured} neural network architecture, which can be combined with lots of advanced algorithms for DRL.

\section{AgentGraph: Structured Deep Reinforcement Learning}
\label{sec:agentgraph}

In this section, we will introduce the proposed structured DRL framework, AgentGraph,  which is based on graph neural networks (GNN). Note that the structured DRL is based on a novel structured neural architecture. It is complementary to various DRL algorithms. Here, we adopt Deep $Q$-Networks (DQN). Next we will first give the background of DQN and GNN, then introduce the proposed structured DRL.

\subsection{Deep-$Q$-Networks (DQN)}

DQN is the first DRL-based algorithm successfully applied in Atari games \cite{Mnih:2013wp}, and then is investigated for dialogue policy optimization. It uses a multi-layer neural network to approximate $Q$-function, $Q_{\theta}(\mathbf{b}_t,a_t)$, i.e. it takes belief state $\mathbf{b}_t$ as input, and predicts the $Q$-values for each action. Compared with the traditional $Q$-learning algorithm \cite{lin-phd1993}, it has two innovations: experience replay and the use of a target network. These techniques help to overcome the instability during the training \cite{mnih-nature2015}.

At every dialogue turn, the agent's experience $\tau=(\mathbf{b}_t,a_t,r_t,\mathbf{b}_{t+1})$ is stored in a pool $\mathcal{D}$. During learning, a batch of experiences are randomly drawn from the pool, i.e.  $\tau \sim {\rm U}(\mathcal{D})$, then Q-learning update rule is applied to update the parameters $\theta$ of $Q$-network:
\begin{equation}
\mathcal{L}(\theta) =  \mathbb{E}_{\tau \sim {\rm U}(\mathcal{D})} \left[\left( y_t   
- Q_{\theta}(\mathbf{b}_t, a_t) \right)^2 \right],
\label{eq:dqn}
\end{equation}
where $y_t = r_t + \gamma \max_{a'} Q_{\hat{\theta}}(\mathbf{b}_{t+1},a')$. Note that the computation of $y_t$ is based on another neural network $Q_{\hat{\theta}}(\mathbf{b}_t,a_t)$, which is referred to as a target network. It is  similar to $Q$-network except that its parameters $\hat{\theta}$ are copied from $\theta$ every $\beta$ steps, and are held fixed during all the other steps.

DQN has many variations. They can be divided into two main categories: One is designing improved RL algorithms for optimizing DQN and another is incorporating new neural network architectures into $Q$-Networks. For example, Double DQN
 (DDQN) \cite{van2016deep} addresses overoptimistic value estimates by decoupling evaluation and selection of an action. Combined with DDQN, prioritized experience replay \cite{schaul2015prioritized} further improves DQN algorithms. The key idea is replaying more often transitions which have high absolute TD-errors. This improves data efficiency and leads to better final policy. In contrast to these improved DQN algorithms, the Dueling DQN \cite{wang2016dueling} changes the network architecture to estimate $Q$-function using separate network heads of value function estimator  $V_{\theta_{1}}(\mathbf{b}_t)$ and advantage function estimator $A_{\theta_{2}}(\mathbf{b}_t,a_t)$:
\begin{equation}
Q_{\theta}(\mathbf{b}_t,a_t) = V_{\theta_1}(\mathbf{b}_t) + A_{\theta_2}(\mathbf{b}_t,a_t).
\label{eq:dueling-dqn-raw}
\end{equation}
The dueling decomposition helps to generalize across actions.

Note that the decomposition in Equation (\ref{eq:dueling-dqn-raw}) doesn't ensure that given $Q$-function $Q_{\theta}(\mathbf{b}_t,a_t)$ we can recover $V_{\theta_1}(\mathbf{b}_t)$ and $A_{\theta_2}(\mathbf{b}_t,a_t)$ uniquely, i.e. we can't conclude that $V_{\theta_1}(\mathbf{b}_t) = \max_{a'}Q_{\theta}(\mathbf{b}_t,a')$. 
To address this problem, the advantage function estimator can be subtracted its maximal value to force it to have zero advantage at the best action:
\begin{equation}
Q_{\theta}(\mathbf{b}_t,a_t) = V_{\theta_1}(\mathbf{b}_t) + \left(  A_{\theta_2}(\mathbf{b}_t,a_t) - \max_{a'} A_{\theta_2}(\mathbf{b}_t,a') \right).
\label{eq:dueling}
\end{equation}
Now we can obtain that  $V_{\theta_1}(\mathbf{b}_t) = \max_{a'}Q_{\theta}(\mathbf{b}_t,a')$.

Similar to Dueling DQN, our proposed AgentGraph is also the innovation of neural network architecture, which is based on graph neural networks.

\subsection{Graph Neural Networks (GNN)}
\label{sec:gnn-intro}

\begin{figure}
\centering
\includegraphics[width=0.45\textwidth]{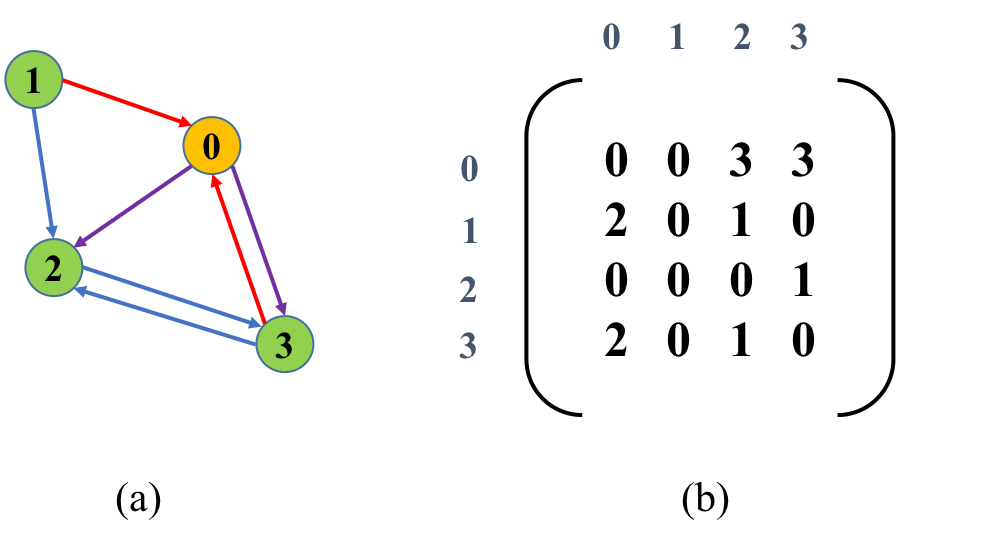}
\caption{(a) An example of a directed graph $G$ with 4 nodes and 7 edges. There are two types of nodes: Nodes 1$\sim$3 (green) are one type of nodes while node 0 (orange) is another. Accordingly, there are 3 types of edges: green $\rightarrow$ green, green $\rightarrow$ orange, and orange $\rightarrow$ green. (b) The adjacency matrix of $G$. 0 denotes that there are no edges between two nodes. 1, 2 and 3 denote three different edge types.}
\label{fig:graph}
\end{figure}

\begin{figure*}
\centering
\includegraphics[width=0.82\textwidth]{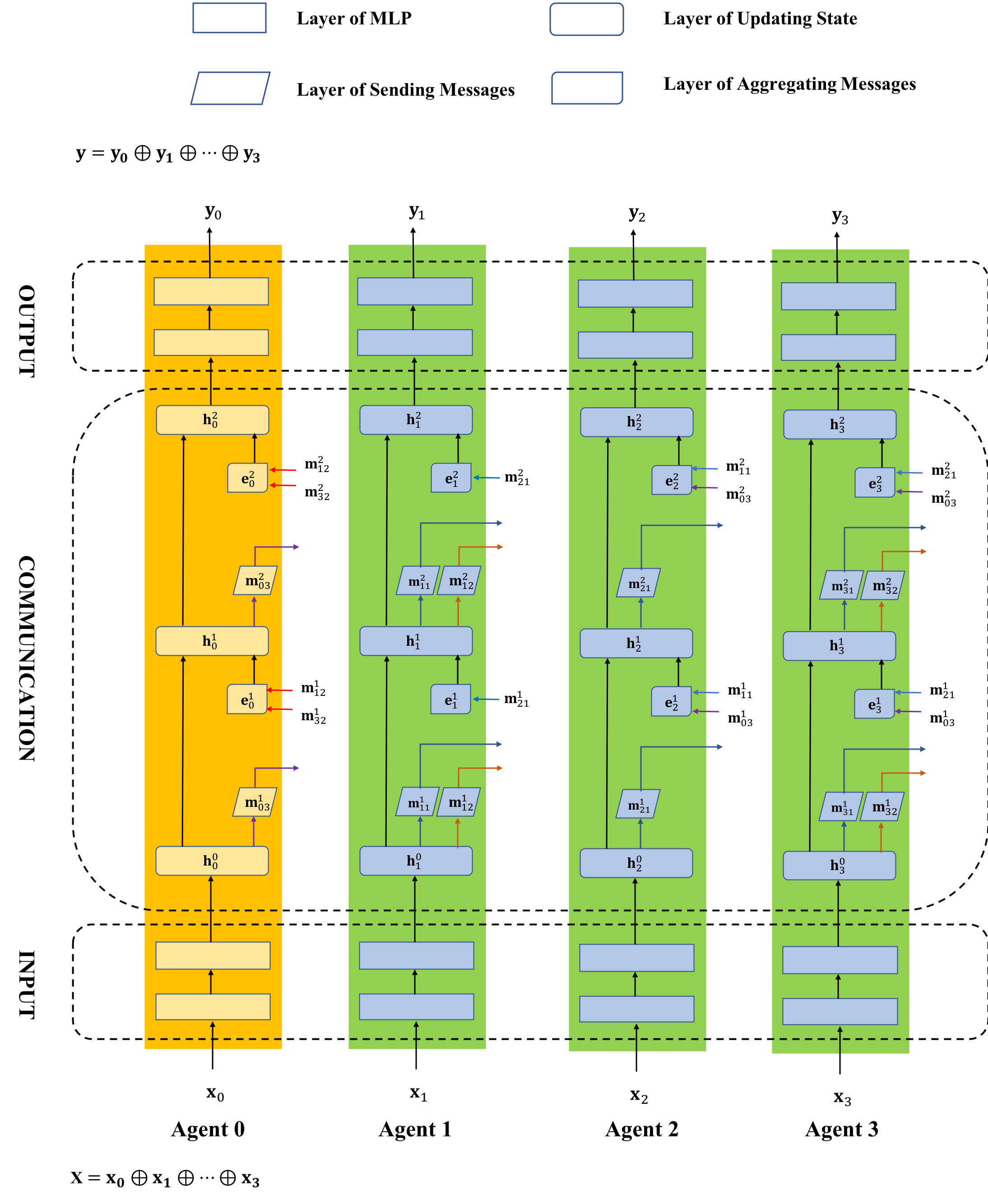}
\vspace{-0.3cm}
\caption{ An illustration of the graph neural network (GNN) according to the graph in Fig. \ref{fig:graph}(a). It consists of 3 parts: input module, communication module, and output module. Here the input module and the output module are both MLP with two hidden layers. The communication module has two communication steps, each one with three operations: sending messages, aggregating messages, and updating state. $\oplus$ denotes concatenation of vectors.}
\label{fig:gnn}
\end{figure*}

We first give some notations before describing the details of GNN. We denote the graph as $G = (V, E)$， where $V$  and $E$ are the set of nodes  $v_i$ and the set of directed edges $e_{ij}$ respectively. $\mathcal{N}_{in}(v_i)$  and $\mathcal{N}_{out}(v_i)$ denote in-coming and out-going neighbors of node $v_i$.
$\mathbf{Z}$ is the adjacency matrix of $G$. The element $z_{ij}$ of $\mathbf{Z}$ is 0 if and only if there is no directed edge from $v_i$ to $v_j$, otherwise $z_{ij}>0$. 
Each node $v_i$ and each edge $e_{ij}$ have an associated node type $c_i$ and an edge type $u_e$ respectively. The edge type is determined by node types. Two edges have the same type if and only if their starting node type and their ending node type both are the same. 
Fig. \ref{fig:graph}(a) shows an example of directed graph $G$ with 4 nodes and 7 edges. Nodes 1$\sim$3 (green) are one type of nodes and node 0 (orange) is another type of node. Accordingly, there are three types of edges in the graph.  Fig. \ref{fig:graph}(b) is the corresponding adjacency matrix.

GNN is a deep neural network associated with the graph $G$ \cite{scarselli2009graph}.  As shown in Fig. \ref{fig:gnn},  it consists of 3 parts: input module, communication module and output module.

\subsubsection{Input Module}
The input $\mathbf{x}$ is divided into some disjoint sub-inputs, i.e. $\mathbf{x} = \mathbf{x}_0 \oplus \mathbf{x}_1 \cdots \oplus \mathbf{x}_n$. Each node (or agent) $v_i (0\leq i \leq n)$ will receive a sub-input $\mathbf{x}_i$, which goes through an input module to obtain a state vector $\mathbf{h}_i^0$ as follows:
\begin{equation}
\mathbf{h}^0_i = f_{c_i}(\mathbf{x}_i),
\end{equation}
where $f_{c_i}$ is an input function for node type $c_i$. For example, in Fig.  \ref{fig:gnn}, it is a multi-layer perceptron (MLP) with two hidden layers.

\subsubsection{Communication Module}
\label{sec:comm}
The communication module takes $\mathbf{h}^0_i$ as the initial state for node $v_i$, then update state from one step (or layer) to the next with following operations. 

\noindent {\bf Sending Messages} At $l$-th step, each agent $v_i$ will send a message $\mathbf{m}_{iu_e}^l$ to its every out-going neighbor $v_j \in \mathcal{N}_{out}(v_i)$:
\begin{equation}
\mathbf{m}_{iu_e}^l = m_{u_e}^l(\mathbf{h}_i^{l-1}),
\end{equation}
where $m_{u_e}^l$ is a function for edge type $u_e$ at $l$-th step. For simplicity, here  a linear transformation $m_{u_e}^l$  is used: $m_{u_e}^l(\mathbf{h}_i^{l-1}) = \mathbf{W}_{u_e}^l \mathbf{h}_i^{l-1}$ , where $\mathbf{W}_{u_e}^l$ is a weight matrix for optimization. It is notable that for the same type of out-going neighbors, the messages sent are the same.

In Fig. \ref{fig:gnn}, there are two communication steps. At the first step, Agent 0 sends message $\mathbf{m}_{03}^1$ to its out-going neighbors Agent 2 and Agent 3. Agent 1 sends messages $\mathbf{m}_{11}^1$ and $\mathbf{m}_{12}^1$ to its two different types of out-going neighbors Agent 2 and Agent 0, respectively. Agent 2 sends message $\mathbf{m}_{21}^1$ to its out-going neighbor Agent 3. Similar to Agent 1, Agent 3 sends messages $\mathbf{m}_{31}^1$ and $\mathbf{m}_{32}^1$ to its two out-going neighbors Agent 2 and Agent 0, respectively.

\noindent {\bf Aggregating Messages} After sending messages, each agent $v_j$ will aggregate messages from its in-coming neighbors, 
\begin{equation}
\begin{split}
\mathbf{e}_j^l & = a^l_{c_j}(\{ \mathbf{m}_{iu_e}^l | v_i \in \mathcal{N}_{in}(v_j)\}), \\
\end{split}
\end{equation}
where the function $a_{c_j}^l$ is the aggregation function for node type $c_j$,  which may be a mean pooling (\textit{Mean-Comm}) or max pooling (\textit{Max-Comm}) function. 

For example, in Fig. \ref{fig:gnn}, at the first communication step, Agent 0 aggregates messages $\mathbf{m}_{12}^1$ and $\mathbf{m}_{32}^1$ from its in-coming neighbors Agent 1 and Agent 3.

\noindent {\bf Updating State} After aggregating messages from neighbors, every agent $v_i$ will update its state from $\mathbf{h}_i^{l-1}$ to $\mathbf{h}_i^l$,

\begin{equation}
\mathbf{h}_i^l = g_{c_i}^l(\mathbf{h}_i^{l-1},\mathbf{e}_i^l),
\end{equation}
where $g_{c_i}^l$ is the update function for node type $c_i$ at $l$-th step, which in practice may be a non-linear layer:  
\begin{equation}
\mathbf{h}_i^l = \sigma(\mathbf{W}_{c_i}^l \mathbf{h}_i^{l-1} + \mathbf{e}_i^l),
\end{equation}
where $\sigma$ is an activation function, e.g. Rectified Linear Unit (ReLU), and $\mathbf{W}_{c_i}^l$ is the transition matrix to be learned.

\subsubsection{Output Module}
After updating state $L$ steps, based on the last state $\mathbf{h}_i^L$ each agent $v_i$ will get the output $\mathbf{y}_i$: 
\begin{equation}
\mathbf{y}_i = o_{c_i} (\mathbf{h}_i^L),
\end{equation}
where $o_{c_i}$ is a function for node type $c_i$, which may be a MLP as shown in Fig. \ref{fig:gnn}.
The final output is the concatenation of all outputs, i.e. $\mathbf{y} = \mathbf{y}_0 \oplus \mathbf{y}_1 \oplus \cdots \oplus \mathbf{y}_n$.

\subsection{Structured DRL}
\label{sec:sdrl}

Combining GNN with DQN, we can obtain structured DQN, which is one of the structured DRL methods. Note that GNN also can be combined with other DRL algorithms, e.g. REINFORCE, A2C. However, in this paper, we focus on DQN algorithm. 

As long as the state and action space can be structured decomposed, the traditional DRL can be replaced by structured DRL. In the next section, we will introduce dialogue policy optimization with structured DRL.

\section{Structured DRL for Dialogue Policy}
\label{sec:sdp}

\begin{figure}[h]
\centering
\includegraphics[width=0.45\textwidth]{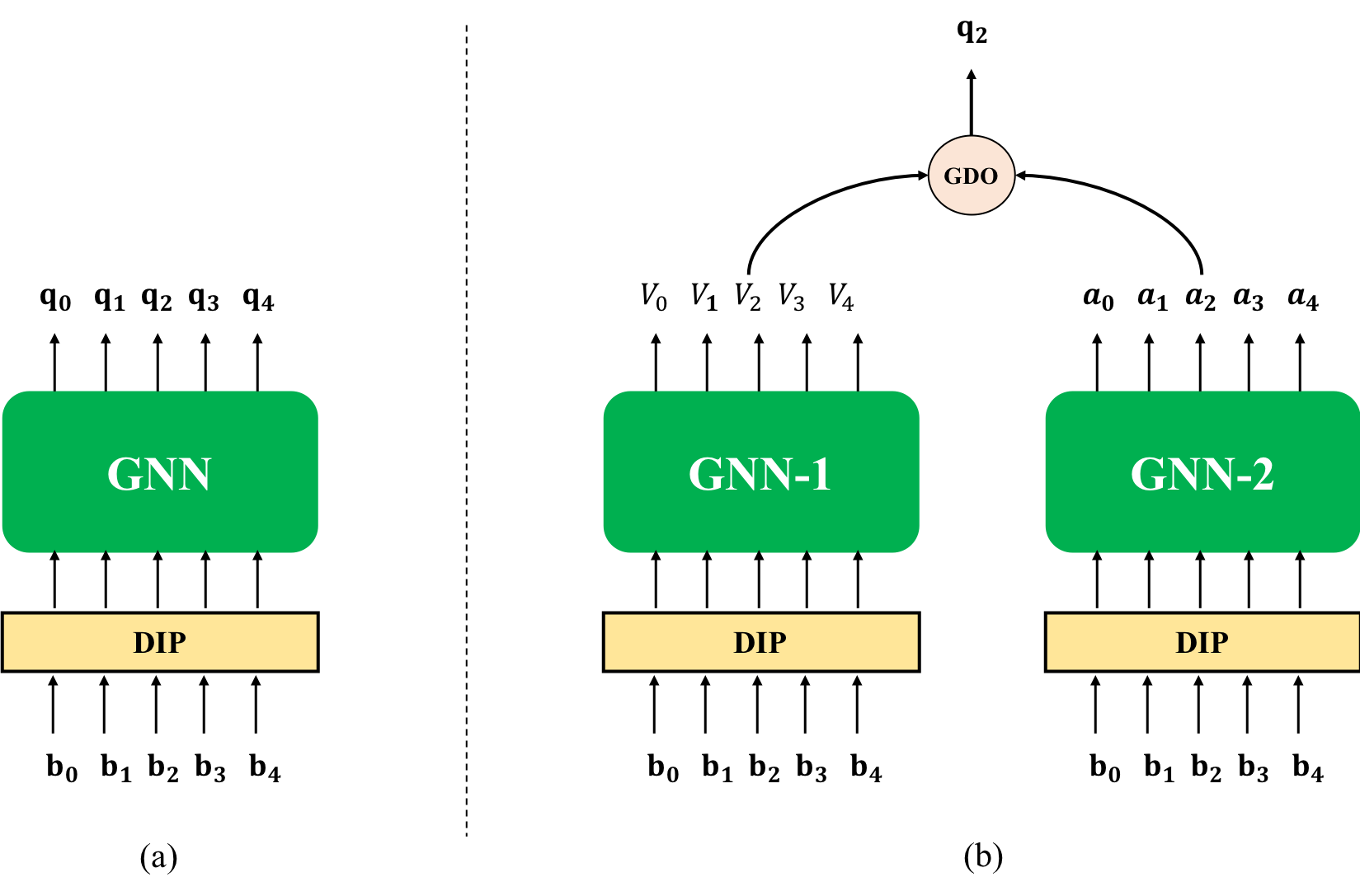}
\caption{(a) An illustration of GNN-based dialogue policy with 5 agents. Agent 0 is slot-independent agent (I-agent), and Agent 1 $\sim$ Agent 4 are slot-dependent agents (S-agents), each one for a slot. More details about the architecture of GNN, please refer to Fig. \ref{fig:gnn}. (b) Dual GNN (DGNN)-based  dialogue policy. It has two GNNs: GNN-1 and GNN-2, each with 5 agents. GDO represents Graph Dueling Operation described in Equation (\ref{eq:gnn-dueling}). $V_i$ is shorthand for $V(\mathbf{b}_i)$, which is a scalar. $\mathbf{a}_i$ is a vector of the advantage values, i.e. $\mathbf{a}_i = [A(\mathbf{b}_i,a^{1}_i), \cdots, A(\mathbf{b}_i,a^{m_i}_i)]$. Note that the GDO here represents the operation of $V_i$ and each element of $\mathbf{a}_i$.}
\label{fig:graph_policy}
\end{figure}

In this section, we introduce two structured DRL methods for dialogue policy optimization: GNN-based dialogue policy and its variant Dual GNN-based dialogue policy. We also discuss three typical graph structures for dialogue policy in section \ref{sec:graph_structure}.

\subsection{Dialogue Policy with GNN}
\label{sec:gnn-policy}

As discussed in section \ref{sec:dm}, the belief dialogue state $\mathbf{b}$\footnote{Note that the subscript $t$ is omitted.} and the set of system actions $\mathbb{A}$ usually can be decomposed, i.e. $\mathbf{b}=\mathbf{b}_{0} \oplus \mathbf{b}_{1}\oplus \cdots \oplus\mathbf{b}_{n}$ and $\mathbb{A} = \mathbb{A}_0 \cup \mathbb{A}_1 \cup \cdots \cup \mathbb{A}_n$.
Therefore, we can design a graph $G$ with $n+1$ nodes for dialogue policy, 
in which there are two types of nodes: a {\em slot-independent} node (I-node) and  $n$ {\em slot-dependent} nodes (S-nodes). Each S-node corresponds to a  slot in the dialogue ontology, while I-node is responsible for slot-independent aspects. The connections between nodes, i.e. the edges of $G$, will be discussed later in subsection \ref{sec:graph_structure}.

The slot-independent belief state $\mathbf{b}_0$ can be used as the input of the I-node, and the marginal belief dialogue state $\mathbf{b}_i$ of $i$-th slot can be used as the input of the $i$-th S-node. However, in practice different slots usually have different number of candidate values, therefore the dimensions of the belief states for two S-nodes are different. In order to abstract the belief state into a fixed size representation, here we use Domain Independent Parametrization (DIP) function $\phi_{dip}^S(\mathbf{b}_i, slot_i)$ \cite{wang2015learning}.
For each slot, $\phi_{dip}^S(\mathbf{b}_i,slot_i)$ generates a summarised
representation of the belief state of the slot $slot_i$. The features can be decomposed into two parts, i.e.
\begin{equation}
    \phi_{dip}^S(\mathbf{b}_i,slot_i) = \phi_1(\mathbf{b}_i) \oplus \phi_2(slot_i),
\end{equation}
where $\phi_1(\mathbf{b}_i)$ represents the summarised \textit{dynamic} features of belief state $\mathbf{b}_i$, including the top three beliefs in $\mathbf{b}_i$, the belief of ``\textit{none}'' value\footnote{For every slot, \textit{``none''} is a special value. It represents that no candidate value of slot has been mentioned by the user.}, the difference between top and second beliefs, the entropy of $slot_i$ and so on. Note that all above features are affected by the output of dialogue state tracker at each turn. 
$\phi_2(slot_i)$ denotes the summarised \textit{static} features of slot $slot_i$. It includes slot length, entropy of the distribution of values of $slot_i$ in the database. These static features represent different characteristics of slots.
They are not affected by the output of dialogue state tracker.

Similarly, another DIP function $\phi_{dip}^I(\mathbf{b}_0)$ is used to extract slot-independent features. It includes last user dialogue act, database search method, whether \textit{offer} has happened and so on.

The architecture of GNN-based dialogue policy is shown in Fig. \ref{fig:graph_policy}(a). The original belief dialogue state is prepossessed with DIP function.
The resulted features with fixed size representation are used as the input of agents in GNN. As discussed in section \ref{sec:gnn-intro}, they are then processed by the input module, communication module and output module.
The output of the I-agent is the Q-values $\mathbf{q}_0$ for the slot-independent actions, i.e. $\mathbf{q}_0 = [Q(\mathbf{b}_0,a^{1}_0),\cdots,Q(\mathbf{b}_0,a^{m_0}_0)]$, where $a_0^j (1\le j \le m_0) \in \mathbb{A}_0$ and $m_0 = |\mathbb{A}_0|$. The output of the $i$-th S-agent is the Q-values $\mathbf{q}_i$ for actions corresponding to $i$-th slot, i.e. $\mathbf{q}_i = [Q(\mathbf{b}_i,a_{i}^1),\cdots,Q(\mathbf{b}_i,a^{m_i}_i)]$, where $a_i^j (1\le j \le m_i) \in \mathbb{A}_i$ and $m_i = |\mathbb{A}_i|$. When making decision, all Q-values are first concatenated, i.e. $\mathbf{q}=\mathbf{q}_{0}\oplus \mathbf{q}_{1}\oplus \cdots \oplus \mathbf{q}_{n}$, then the action is chosen according to $\mathbf{q}$ as done in vanilla DQN.

Compared with traditional DRL-based dialogue policy, the GNN-based policy has some advantages: First, due to the use of DIP features and the benefit of GNN architecture, S-agents share all parameters. With these shared parameters, the skills can be transferred between S-agents, which can improve the speed of learning. Moreover,  when a new domain exists, the policy trained in another domain can be used to initialize the policy in the new domain\footnote{We will discuss policy transfer learning with AgentGraph in section \ref{sec:policy_transfer}.}.

\subsection{Dialogue Policy with Dual GNN (DGNN)}

As introduced in the previous subsection, although GNN-based dialogue policy utilizes structured architecture of the network, it conducts \textit{flat} decision as traditional DRL does. It's shown that flat RL suffers from scalability to domains with a large number of slots \cite{baolin-hier}.
In contrast, hierarchical RL decomposes the decisions in several steps and uses different abstraction levels in each sub-decision. This \textit{hierarchical} decision procedure makes it well suited to large dialogue domains.

Recently proposed Feudal Dialogue Management (FDM) \cite{casanueva2018feudal} is a typical hierarchical method, in which there are three types of policies: a master policy, a slot-independent policy, and a set of slot-dependent policies, one for each slot. At each turn, the master policy first decides to take either a slot-independent or slot-dependent action. Then the corresponding slot-independent policy or slot-dependent policies are used to choose a primitive action.
During the training phase, each type of dialogue policy has its private replay memory, and their parameters are updated independently. 

Inspired by FDM and Dueling DQN, here we propose a \textit{differentiable}  end-to-end hierarchical framework, Dual GNN (DGNN)-based dialogue policy. As shown in Fig. \ref{fig:graph_policy}(b), there are two streams of GNNs. 
One (GNN-1) is to estimate the value function $V(\mathbf{b}_i)$ for each agent. The architecture of GNN-1 is similar to that of GNN-based dialogue policy in Fig. \ref{fig:graph_structure}(a) except that the dimension of output for each agent is 1. The output $V(\mathbf{b}_i)$ represents the expected discounted cumulative return when selecting the best action from $\mathbb{A}_i$ at the the belief state $\mathbf{b}$, i.e. $V(\mathbf{b}_i) = \max_{a^{j}_i \in \mathbb{A}_i} Q(\mathbf{b}_i,a^{j}_i)$.
GNN-2 is to estimate the advantage function $A(\mathbf{b}_i,a^{j}_i)$ of choosing $j$-th action in $i$-th agent. The architecture of GNN-2 is same as that of GNN-based dialogue policy.
With the value function and the advantage function, for each agent, the $Q$-function $Q(\mathbf{b}_i,a^{j}_i)$ can be written as
\begin{equation}
Q(\mathbf{b}_i,a^{j}_i) = V(\mathbf{b}_i) + A(\mathbf{b}_i,a^{j}_i).
\label{eq:gnn-dueling-raw}
\end{equation}
Similar to Equation (\ref{eq:dueling}), in order to make sure that $V(\mathbf{b}_i)$ and $A(\mathbf{b}_i,a^{j}_i)$ are appropriate value function estimator and advantage function estimator, Equation (\ref{eq:gnn-dueling-raw}) can be reformulated as 
\begin{equation}
Q(\mathbf{b}_i,a^{j}_i) = V(\mathbf{b}_i) + \left(  A(\mathbf{b}_i,a^{j}_i) - \max_{a'\in \mathbb{A}_i} A(\mathbf{b}_i,a')  \right).
\label{eq:gnn-dueling}
\end{equation}
This is called \textit{Graph Dueling Operation} (GDO).
With GDO, the parameters of two GNNs (GNN-1 and GNN-2) can be jointly trained. 

Compared with GNN-based policy and Feudal policy, DGNN-based dialogue policy integrates two-level decisions into a single decision with GDO at each turn.  
GNN-1 is \textit{implicitly} to make a high-level decision choosing an agent to select primitive action. GNN-2 is \textit{implicitly} to make a low-level decision choosing a primitive action from the previously selected agent.

\subsection{Three Graph Structures for Dialogue Policy}
\label{sec:graph_structure}

\begin{figure}[h]
\centering
\includegraphics[width=0.45\textwidth]{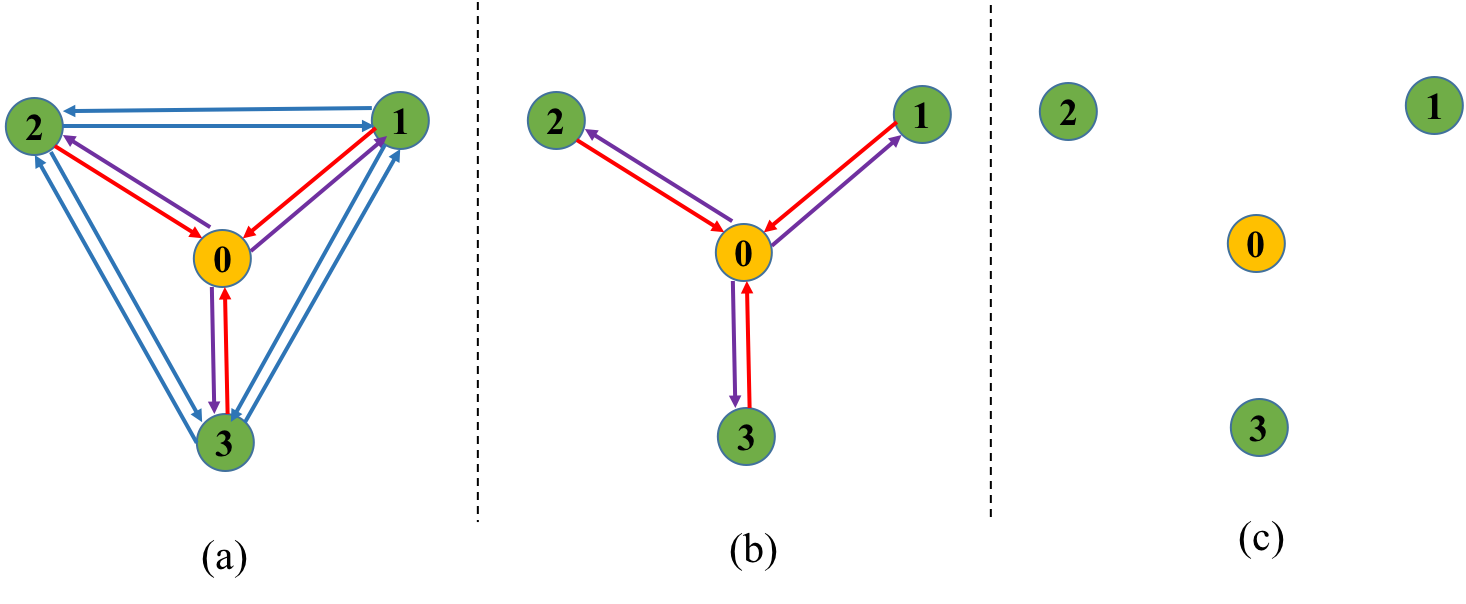}
\vspace{-0.4cm}
\caption{Three different graph structures. (a) FC: a fully connected graph. (b) MN: a master-node graph. (c) FU: an isolated graph.}
\label{fig:graph_structure}
\end{figure}

In previous sections, we assume that the structure of graph $G$, i.e. the adjacency matrix $\mathbf{Z}$, is known. However, in practice, the relations between slots are usually not well defined. Therefore the graph is not known. Here we investigate three different graph structures of GNN: {\bf FC}, {\bf MN} and {\bf FU}.
\begin{itemize}
\item {\bf FC}: a {\em fully} connected graph as shown in Fig. \ref{fig:graph_structure}(a), i.e. there are two directed edges between every two nodes. As discussed in previous sections, it has three types of edges: S-node $\to$ S-node, S-node $\to$ I-node and I-node $\to$ S-node.

\item {\bf MN}: a {\em master}-node graph as shown in Fig. \ref{fig:graph_structure}(b). The I-node is the master node during the communication, which means there are edges only between the I-node and the S-nodes and no edges between the S-nodes.

\item {\bf FU}: an {\em isolated} graph as shown in Fig. \ref{fig:graph_structure}(c). There is no edges between every two nodes.

\end{itemize}

Note that DGNN-based dialogue policy has two GNNs, GNN-1 and GNN-2. For GNN-1, it determines from which node final action is selected. It is a \textit{global} decision, and the communication between nodes is necessary. However, for GNN-2, it determines which action is to be selected in each node. It is a \textit{local} decision, and there is no need to exchange messages between nodes. Therefore, in this paper, we only compare different structures of GNN-1 and use FU as the graph structure of GNN-2.

\section{Dialogue Policy Transfer Learning}
\label{sec:policy_transfer}

In the real-world scenario where the conversation agent directly interacts with users, the performance at the early training period is very important. 
The policy trained from scratch is usually rather poor in the early stages of learning, which may result in bad user experience and hence it is hard to attract enough users to have more interactions for further policy training. 
This is called \textit{safety} problem of online dialogue policy learning\cite{chen-eacl2017,chen-emnlp2017,chang2017affordable}. 

Policy adaptation is one way to solve this problem \cite{chen2017policy}. 
However, for traditional DRL-based dialogue policy, it's still challenging for policy transfer between different domains, because the ontologies of the two domains are different, as a result, the action sets and the state spaces both are fundamentally different. 
Our proposed AgentGraph-based policy can be directly transferred from one domain to another domain. As introduced in the previous section, AgentGraph has an I-agent and $n$ S-agent, each one for a slot. All S-agents share parameters. Even though slots between the target domain and the source domain are different, the shared parameters of S-agent and the parameters of I-agent can be used to initialize the parameters of AgentGraph in the target domain.

\section{Experiments}
\label{sec:exp}

In this section, we evaluate the performance of our proposed AgentGraph methods. Section \ref{sec:exp-setup} introduces the set-up of evaluation. In section \ref{sec:exp-performance} we compare the performance of AgentGraph methods with traditional RL methods.
Section \ref{sec:exp-structure}  investigates the effect of graph structures and communication methods. In section \ref{sec:exp-transfer}, we examine the transfer of dialogue policy with AgentGraph model.

\subsection{Evaluation Set-up}
\label{sec:exp-setup}

\begin{table}[h]
\begin{center}
\caption{The set of benchmark environments}
\begin{tabular}{|c|c|c|c|}
\hline
Environment & SER & Masks & User \\
\hline
Env.1 & 0\% & Yes & Standard \\
Env.2 & 0\% & No & Standard \\
Env.3 & 15\% & Yes & Standard \\
Env.4 & 15\% & No & Standard \\
Env.5 & 15\% & Yes & Unfriendly \\
Env.6 & 30\% & Yes & Standard \\
\hline
\end{tabular}
\label{tab:env}
\end{center}
\end{table}
\subsubsection{PyDial Benchmark}

RL-based dialogue policies are typically evaluated on a small set of simulated or crowd-sourcing environments. It is difficult to perform a fair comparison between different models, because these environments are built by different research groups and are not available to the community.
Fortunately, a common benchmark is recently published in \cite{casanueva2017benchmarking}, which can evaluate the capability of policies in extensive simulated environments. 
These environments are implemented based on an open-source toolkit: PyDial \cite{ultes2017pydial}, which is a multi-domain SDS toolkit with domain-independent implementations of all the SDS modules, simulated users and error models.
There are 6 environments across a number of dimensions in the benchmark, which will be briefly introduced next and are summarized in Table \ref{tab:env}.

The first dimension of variability is the \textbf{semantic error rate (SER)}, which simulates different noise levels in the input module of SDS. Here SER is set to three different values, 0\%, 15\% and 30\%.

The second dimension of variability is the \textbf{user model}. Env.5's user model is defined to be an \textit{Unfriendly} distribution, where the users barely provide any extra information to the system. The  others' are all \textit{Standard}.

The last dimension of variability comes from the action \textbf{masking mechanism}. In practice, some heuristics are usually used to mask the invalid actions when making a decision. For example, the action \textit{confirm}($slot_i$) is masked if all the probability mass of $slot_i$ is in the ``{\em none}" value. Here in order to evaluate the learning capability of the models, the action masking mechanism is disabled in two of the environments: Env.2 and Env.4. 

In addition, there are three different \textbf{domains}: information seeking tasks for restaurants in Cambridge (CR) and San Francisco (SFR) and a generic shopping task for laptops (LAP). They are slot-based, which means the dialogue state is factorized into slots. CR, SFR and LAP have 3, 6 and 11 slots respectively. Usually, more slots have, the task is more difficult.

In total, there are 18 tasks\footnote{We use ``Domain-Environment'' to represent each task, e.g. SFR-Env.1 represents the domain SFR in the Env.1.} with 6 environments and 3 domains in PyDial benchmark. We will evaluate our proposed methods on all these tasks.

\subsubsection{Models}
There are 5 different AgentGraph models for evaluation.
\begin{itemize}
    \item \textbf{FM-GNN}: GNN-based dialogue policy with {\em fully} connected (FC) graph. The communication method between nodes is \textit{Mean-Comm}. 
    \item \textbf{FM-DGNN}: DGNN-based dialogue policy with {\em fully} connected (FC) graph. The communication method between nodes is \textit{Mean-Comm}. 
    \item \textbf{UM-DGNN}: It is similar to FM-DGNN except that the {\em isolated} (FU) graph is used.
    \item \textbf{MM-DGNN}: It is similar to FM-DGNN except that the {\em master-node} (MN) graph is used.
    \item \textbf{FX-DGNN}: It is similar to FM-DGNN except that the communication method is {\em Max-Comm}. 
\end{itemize}
These models are summarised in Table \ref{tab:models}.

For both GNN-based and DGNN-based models, the inputs of S-agents and I-agent are 25 DIP features and 74 DIP features respectively. Each S-agent has 3 actions (\textit{request, confirm} and \textit {select}) and the I-agent has 5 actions (\textit{inform by constraints, inform requested, inform alternatives, bye} and \textit{request more}).
More details about DIP features and actions used here, please refer to \cite{casanueva2018feudal} and \cite{casanueva2017benchmarking}. 
We use grid-search to find the best hyper-parameters of GNN/DGNN. 
The input module is one layer MLP. The output dimensions of the input module for I-agent and S-agents are 250 and 40 respectively. 
For the communication module, the number of communication steps (or layers)  $L$ is 1.
The output dimensions of the communication module for I-agent and S-agents are 100 and 20 respectively.
The output module is also one layer MLP. The output dimensions are the corresponding number of actions of S-agents and I-agent. 

\subsubsection{Evaluation Metrics}
For each task, every model is trained over ten different random seeds ($0\sim 9$). After each 200 training dialogues, the models are evaluated over 500 test dialogues and the results shown are averaged over all 10 seeds.

The evaluation metrics used here are the average reward and the average success rate.  The success rate is the percentage of dialogues which are completed successfully. The reward is defined as $20 \times \mathbbm{1}(\mathcal{D}) - T$, here $\mathbbm{1}(\mathcal{D})$ is the success indicator and $T$ is the number of dialogue turns.

\begin{table}
\begin{center}
\caption{Summary of AgentGraph Models}
\begin{tabular}{|c|c|c|c|}
\hline
Models & Dual & Comm. & Graph Structure \\
\hline
FM-GNN & No & Mean & Fully Connected \\
FM-DGNN & Yes & Mean & Fully Connected \\
UM-DGNN & Yes & Mean & Isolated \\
MM-DGNN & Yes & Mean & Master-node \\
FX-DGNN & Yes & Max & Fully Connected \\
\hline
\end{tabular}
\label{tab:models}
\end{center}
\end{table}

\begin{figure*}
\centering
\includegraphics[width=0.90\textwidth]{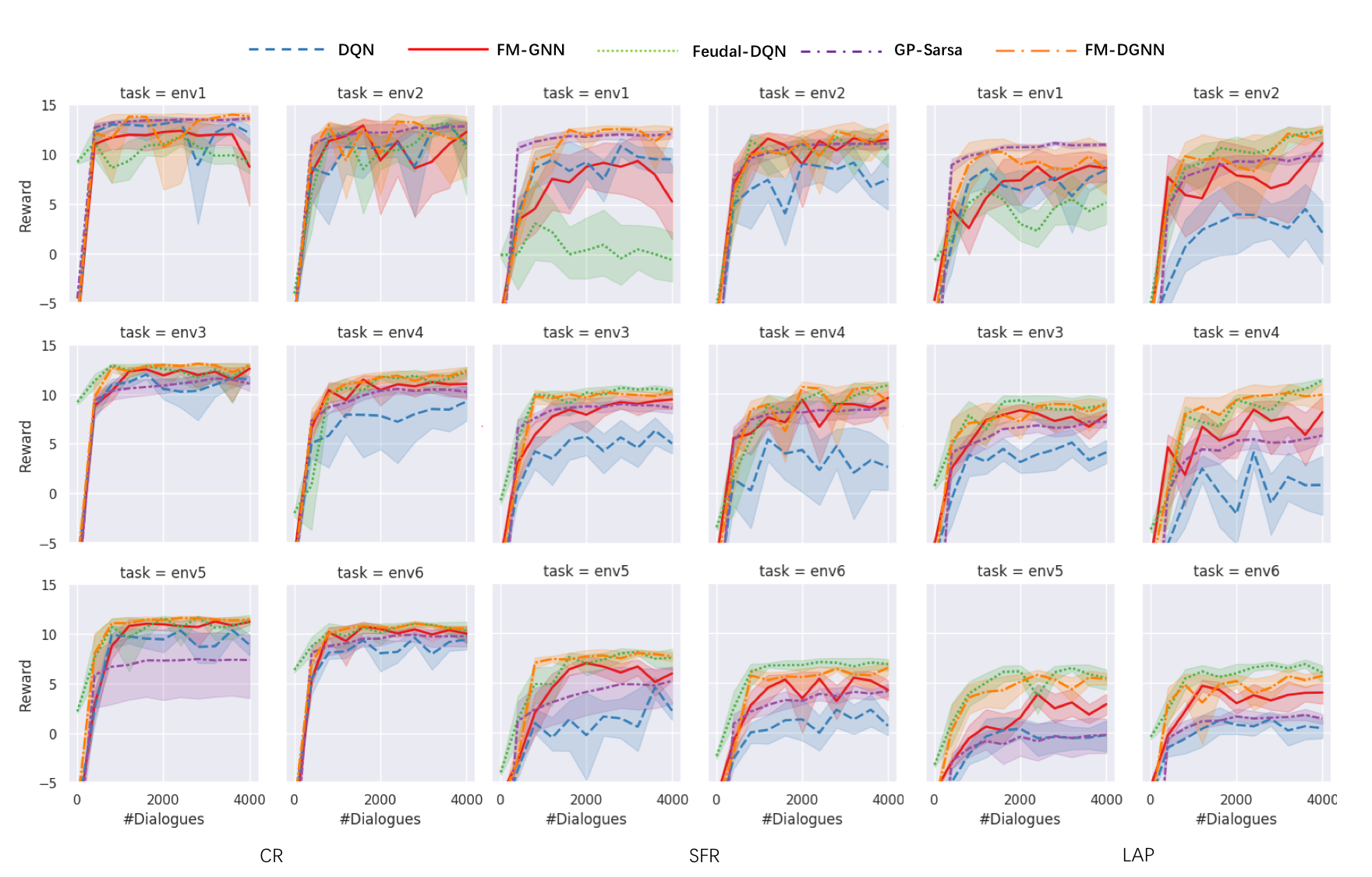}
\vspace{0.5cm}
\includegraphics[width=0.90\textwidth]{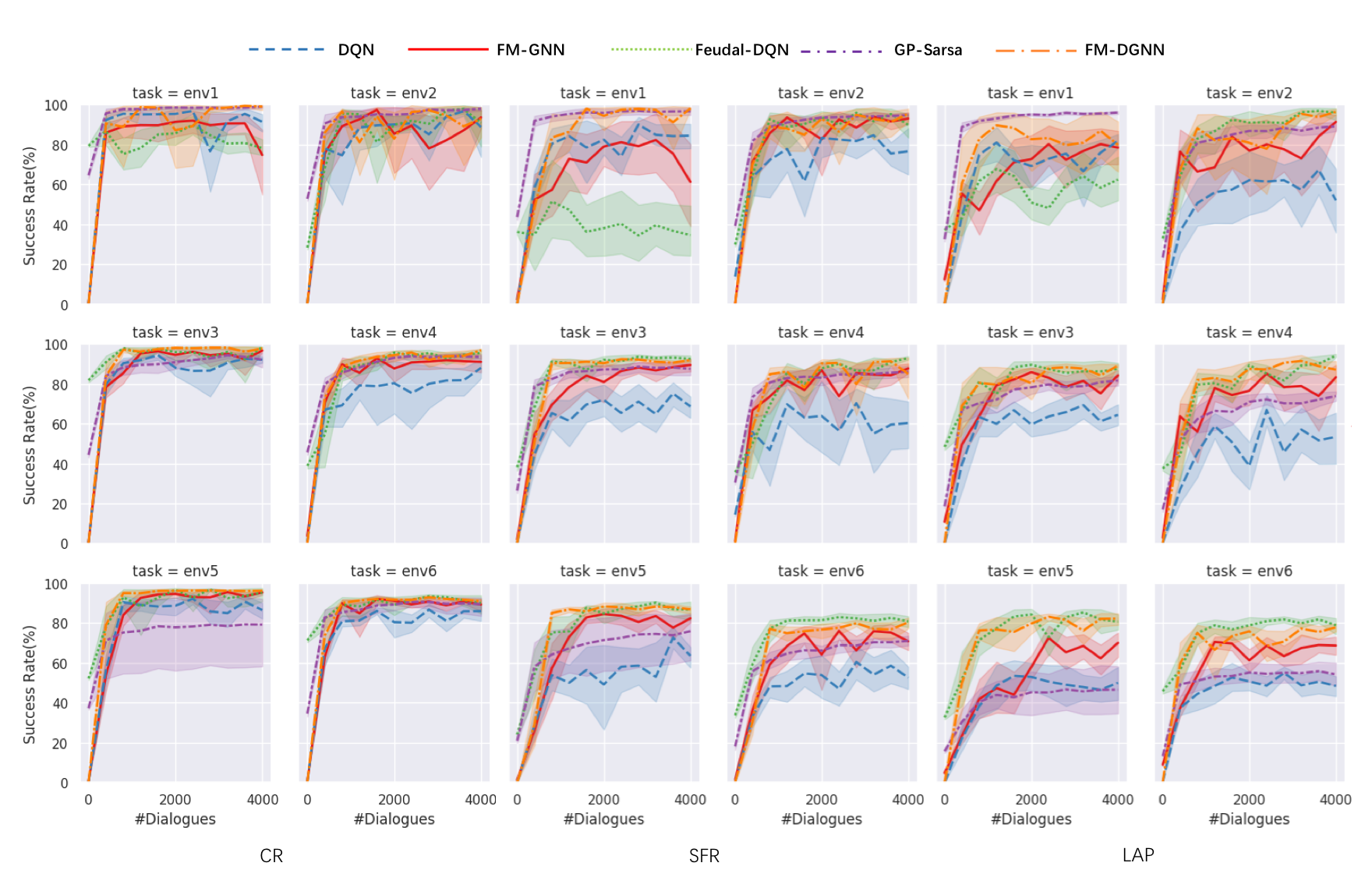}
\caption{The learning curves of reward and success rates for different dialogue policies (GP-Sarsa, DQN, FM-GNN, and FM-DGNN) on 18 different tasks.}
\label{fig:exp-drl}
\end{figure*}

\begin{center}
\begin{table*}[h]
\small
\centering
\caption{Reward and success rates after 1000/4000 training dialogues. The results in bold blue are the best success rates, and the results in bold black are the best rewards.}
\begin{tabular}{rccccccccccccccccl}
\hline
 & & \multicolumn{6}{c}{Baselines} & \multicolumn{10}{c}{Structured DRL} \\
\hline
& & \multicolumn{2}{c}{GP-Sarsa} & \multicolumn{2}{c}{DQN} & \multicolumn{2}{c}{Feudal-DQN} & \multicolumn{2}{c}{FM-GNN} & \multicolumn{2}{c}{UM-DGNN} & \multicolumn{2}{c}{MM-DGNN} & \multicolumn{2}{c}{FX-DGNN} & \multicolumn{2}{c}{FM-DGNN} \\
 \multicolumn{2}{c}{\emph{Task}} & Suc. & Rew. & Suc. & Rew. & Suc. & Rew.& Suc. & Rew.& Suc. & Rew.& Suc. & Rew.& Suc. & Rew.& Suc. & Rew.\\
 \hline
 \multicolumn{16}{c}{ after 1000 training dialogues}  \\
 \hline
 \multirow{3}{*}{\rotatebox{90}{Env.1}}
& CR & 97.8 & 13.3 & 87.8 & 11.4 & 82.4 & 10.4 & 89.7 & 11.9 & 85.7 & 11.0 & 96.8 & 13.1 & 95.8 & 12.6 & {\color{blue} \textbf{98.0}} & \textbf{13.5}\\
& SFR & {\color{blue} \textbf{95.6}} & \textbf{11.6} & 81.5 & 9.0 & 43.6 & 1.6 & 71.1 & 7.2 & 46.3 & 2.2 & 86.0 & 10.2 & 91.0 & 11.0 & 84.8 & 9.5\\
& LAP & 91.6 & 9.9 & 75.6 & 7.4 & 56.0 & 4.0 & 58.6 & 5.0 & 55.7 & 4.1 & 76.5 & 7.6 & {\color{blue} \textbf{93.2}} & \textbf{11.1} & 90.0 & 10.5\\
 \hline
 \multirow{3}{*}{\rotatebox{90}{Env.2}}
& CR & 94.5 & 12.1 & 71.8 & 7.0 & 91.9 & 11.2 & 83.6 & 9.9 & 88.6 & 10.7 & 88.9 & 11.5 & {\color{blue} \textbf{95.3}} & \textbf{12.5} & 88.8 & 11.0\\
& SFR & 90.2 & 10.1 & 77.7 & 7.8 & 89.5 & 10.0 & 81.2 & 8.4 & 83.0 & 9.5 & {\color{blue} \textbf{91.9}} & \textbf{10.9} & 89.3 & 10.8 & 84.5 & 9.8\\
& LAP & 82.4 & 8.1 & 57.7 & 3.0 & 81.5 & 8.6 & 75.9 & 7.7 & 88.4 & 9.8 & {\color{blue} \textbf{94.5}} & \textbf{11.3} & 88.9 & 10.1 & 89.7 & 10.0\\
 \hline
 \multirow{3}{*}{\rotatebox{90}{Env.3}}
& CR & 89.0 & 10.4 & 90.9 & 11.1 & 97.2 & 12.7 & 95.8 & 12.4 & 81.0 & 8.9 & 97.6 & 12.8 & 97.0 & 12.6 & {\color{blue} \textbf{97.9}} & \textbf{13.0}\\
& SFR & 82.2 & 7.6 & 66.1 & 4.6 & 90.6 & \textbf{9.8} & 75.2 & 6.9 & 80.6 & 6.7 & {\color{blue} \textbf{90.8}} & 9.6 & 90.2 & 9.5 & 90.3 & 9.4\\
& LAP & 72.7 & 5.5 & 60.1 & 3.5 & 84.0 & 8.2 & 67.3 & 5.5 & 72.2 & 4.9 & 85.6 & 7.9 & {\color{blue} \textbf{88.7}} & \textbf{8.9} & 79.8 & 7.1\\
 \hline
 \multirow{3}{*}{\rotatebox{90}{Env.4}}
& CR & 87.5 & 9.1 & 79.8 & 8.5 & {\color{blue} \textbf{91.4}} & 10.9 & 84.9 & 9.4 & 83.5 & 8.9 & 91.0 & 10.9 & 90.4 & 10.7 & 91.2 & \textbf{11.0}\\
& SFR & 81.3 & 7.5 & 68.3 & 5.4 & 83.2 & 8.4 & 76.0 & 6.3 & 79.7 & 7.7 & 84.1 & 8.7 & 84.5 & 8.8 & {\color{blue} \textbf{87.8}} & \textbf{9.7}\\
& LAP & 64.6 & 4.2 & 40.9 & -1.3 & 84.9 & 8.6 & 62.8 & 3.5 & 74.8 & 6.0 & 83.8 & 8.4 & {\color{blue} \textbf{86.0}} & \textbf{9.2} & 82.9 & 8.1\\
 \hline
 \multirow{3}{*}{\rotatebox{90}{Env.5}}
& CR & 76.3 & 6.9 & 90.5 & 10.0 & 92.4 & 10.6 & 91.9 & 10.4 & 55.7 & 2.2 & {\color{blue} \textbf{96.0}} & \textbf{11.4} & 94.4 & 10.8 & 95.6 & 11.3\\
& SFR & 66.5 & 2.8 & 55.0 & 1.2 & 82.5 & 6.6 & 62.0 & 3.1 & 26.2 & -4.2 & 85.3 & 6.9 & {\color{blue} \textbf{86.4}} & \textbf{7.3} & 86.1 & 7.2\\
& LAP & 42.1 & -1.2 & 43.6 & -1.5 & 74.7 & 4.2 & 43.5 & -0.2 & 47.1 & -0.6 & 69.4 & 3.0 & {\color{blue} \textbf{81.7}} & \textbf{5.2} & 80.4 & 5.0\\
 \hline
 \multirow{3}{*}{\rotatebox{90}{Env.6}}
& CR & 87.5 & 9.2 & 84.2 & 8.9 & 87.2 & 9.7 & 91.0 & 10.4 & 65.4 & 5.1 & {\color{blue} \textbf{91.9}} & 10.6 & 91.8 & \textbf{10.7} & 91.7 & \textbf{10.7}\\
& SFR & 64.0 & 2.9 & 55.5 & 1.7 & {\color{blue} \textbf{80.2}} & \textbf{6.6} & 68.0 & 4.2 & 63.1 & 2.5 & 74.4 & 5.1 & 66.7 & 3.7 & 73.6 & 4.7\\
& LAP & 54.2 & 1.2 & 46.7 & -0.1 & 74.2 & \textbf{5.2} & 70.5 & 4.6 & 64.6 & 3.0 & {\color{blue} \textbf{75.7}} & \textbf{5.2} & 74.4 & 4.7 & 71.5 & 4.2\\
 \hline
 \multirow{3}{*}{\rotatebox{90}{Mean}}
& CR & 88.8 & 10.2 & 84.2 & 9.5 & 90.4 & 10.9 & 89.5 & 10.7 & 76.6 & 7.8 & 93.7 & 11.7 & {\color{blue} \textbf{94.1}} & 11.6 & 93.9 & \textbf{11.8}\\
& SFR & 80.0 & 7.1 & 67.4 & 4.9 & 78.3 & 7.2 & 72.2 & 6.0 & 63.2 & 4.1 & {\color{blue} \textbf{85.4}} & \textbf{8.6} & 84.7 & 8.5 & 84.5 & 8.4\\
& LAP & 67.9 & 4.6 & 54.1 & 1.8 & 75.9 & 6.5 & 63.1 & 4.4 & 67.1 & 4.5 & 80.9 & 7.2 & {\color{blue} \textbf{85.5}} & \textbf{8.2} & 82.4 & 7.5\\
 \hline
 
 \hline
\multicolumn{16}{c}{ after 4000 training dialogues}  \\
 \hline
 \multirow{3}{*}{\rotatebox{90}{Env.1}}
& CR & 98.9 & 13.6 & 91.4 & 12.1 & 78.2 & 9.2 & 74.8 & 8.7 & 92.1 & 12.3 & 86.9 & 11.2 & {\color{blue} \textbf{99.4}} & \textbf{14.0} & 99.0 & 13.8\\
& SFR & 96.7 & 12.0 & 84.5 & 9.5 & 34.6 & -0.6 & 61.3 & 5.3 & 52.9 & 3.7 & 94.8 & 12.0 & {\color{blue} \textbf{98.1}} & \textbf{12.7} & {\color{blue} \textbf{98.1}} & 12.6\\
& LAP & {\color{blue} \textbf{96.1}} & \textbf{11.0} & 81.8 & 8.5 & 62.5 & 5.2 & 78.5 & 8.6 & 48.6 & 2.9 & 79.1 & 8.0 & 86.0 & 9.2 & 79.8 & 8.5\\
 \hline
 \multirow{3}{*}{\rotatebox{90}{Env.2}}
& CR & {\color{blue} \textbf{97.9}} & 12.8 & 88.7 & 11.0 & 90.8 & 10.5 & 93.6 & 12.2 & 82.0 & 10.0 & 96.3 & \textbf{13.2} & 93.1 & 12.6 & 93.0 & 11.6\\
& SFR & 94.7 & 11.1 & 76.6 & 7.5 & 89.8 & 10.3 & 93.0 & 11.5 & 79.0 & 8.4 & 92.8 & 10.8 & 94.4 & 11.8 & {\color{blue} \textbf{95.4}} & \textbf{12.5}\\
& LAP & 89.1 & 9.9 & 52.0 & 2.2 & 96.0 & 12.1 & 91.4 & 11.1 & 66.4 & 5.8 & 94.1 & 11.7 & 95.4 & 12.0 & {\color{blue} \textbf{96.5}} & \textbf{12.5}\\
 \hline
 \multirow{3}{*}{\rotatebox{90}{Env.3}}
& CR & 92.1 & 11.1 & 92.1 & 11.5 & {\color{blue} \textbf{98.4}} & \textbf{13.0} & 96.6 & 12.6 & 80.6 & 8.7 & 97.6 & 12.9 & 96.5 & 12.6 & 97.7 & 12.9\\
& SFR & 87.5 & 8.6 & 68.6 & 5.0 & {\color{blue} \textbf{92.5}} & 10.2 & 89.4 & 9.4 & 70.9 & 4.9 & 92.3 & \textbf{10.3} & 90.8 & 10.0 & 91.9 & 10.2\\
& LAP & 81.6 & 7.2 & 64.4 & 4.1 & 87.4 & 8.9 & 84.2 & 7.9 & 76.8 & 5.7 & 87.6 & 8.5 & 86.3 & 8.3 & {\color{blue} \textbf{89.1}} & \textbf{9.1}\\
 \hline
 \multirow{3}{*}{\rotatebox{90}{Env.4}}
& CR & 93.4 & 10.2 & 88.0 & 9.3 & 95.5 & 12.3 & 90.9 & 11.0 & 86.4 & 10.0 & 95.8 & 12.1 & 93.4 & 11.4 & {\color{blue} \textbf{96.8}} & \textbf{12.4}\\
& SFR & 85.9 & 8.6 & 60.3 & 2.7 & {\color{blue} \textbf{92.9}} & \textbf{10.8} & 87.7 & 9.6 & 79.3 & 7.8 & 88.4 & 10.2 & 89.2 & 10.3 & 84.9 & 9.2\\
& LAP & 73.8 & 5.8 & 53.4 & 0.8 & {\color{blue} \textbf{94.2}} & \textbf{11.3} & 83.3 & 8.2 & 62.4 & 3.3 & 87.4 & 9.8 & 87.8 & 9.3 & 87.0 & 9.9\\
 \hline
 \multirow{3}{*}{\rotatebox{90}{Env.5}}
& CR & 79.2 & 7.3 & 86.4 & 8.9 & 95.2 & 11.3 & 95.2 & 11.2 & 60.9 & 3.2 & 95.9 & \textbf{11.4} & 95.7 & \textbf{11.4} & {\color{blue} \textbf{96.1}} & \textbf{11.4}\\
& SFR & 75.9 & 5.2 & 63.5 & 2.2 & 86.7 & 7.5 & 82.3 & 5.9 & 62.8 & 1.6 & 86.3 & 7.4 & 86.3 & 7.3 & {\color{blue} \textbf{86.8}} & \textbf{7.7}\\
& LAP & 46.5 & -0.2 & 50.0 & -0.2 & 80.7 & \textbf{5.5} & 70.0 & 2.8 & 56.2 & 0.9 & 79.4 & 5.0 & 77.9 & 4.0 & {\color{blue} \textbf{82.2}} & 5.4\\
 \hline
 \multirow{3}{*}{\rotatebox{90}{Env.6}}
& CR & 89.4 & 9.8 & 85.9 & 9.4 & 89.9 & 10.3 & 89.3 & 10.0 & 71.4 & 6.2 & {\color{blue} \textbf{92.8}} & 10.7 & 92.6 & \textbf{10.9} & 90.9 & 10.5\\
& SFR & 71.0 & 4.2 & 52.5 & 0.7 & {\color{blue} \textbf{80.8}} & \textbf{6.9} & 70.8 & 4.3 & 63.3 & 2.8 & 80.1 & 6.5 & 71.0 & 5.0 & 80.4 & 6.5\\
& LAP & 54.2 & 1.4 & 48.4 & 0.4 & {\color{blue} \textbf{78.8}} & \textbf{6.0} & 68.7 & 4.0 & 62.1 & 2.3 & 67.4 & 3.7 & 76.5 & 5.5 & 77.8 & 5.7\\
 \hline
 \multirow{3}{*}{\rotatebox{90}{Mean}}
& CR & 91.8 & 10.8 & 88.8 & 10.4 & 91.3 & 11.1 & 90.1 & 11.0 & 78.9 & 8.4 & 94.2 & 11.9 & 95.1 & \textbf{12.2} & {\color{blue} \textbf{95.6}} & 12.1\\
& SFR & 85.3 & 8.3 & 67.7 & 4.6 & 79.6 & 7.5 & 80.8 & 7.7 & 68.0 & 4.9 & 89.1 & 9.5 & 88.3 & 9.5 & {\color{blue} \textbf{89.6}} & \textbf{9.8}\\
& LAP & 73.6 & 5.8 & 58.3 & 2.6 & 83.3 & 8.2 & 79.4 & 7.1 & 62.1 & 3.5 & 82.5 & 7.8 & 85.0 & 8.0 & {\color{blue} \textbf{85.4}} & \textbf{8.5}\\
 \hline

\end{tabular}
\label{tab:results}
\end{table*}
\end{center}

\subsection{Performance of AgentGraph Models}
\label{sec:exp-performance}
In this subsection, we compare the proposed AgentGraph models (FM-DGNN and FM-GNN) with traditional RL models (GP-Sarsa and DQN)\footnote{For GP-Sarsa and DQN, we use the default set-up in PyDial Benchmark.}. The learning curves of success rates and reward are shown in Fig. \ref{fig:exp-drl}, and the reward and success rates after 1000 and 4000 training dialogues for these models are summarised in Table \ref{tab:results}\footnote{For the sake of brevity, the standard deviation in the table is omitted. Compared with GP-Sarsa/DQN/Feudal-DQN, FM-DGNN performs significantly better on 12/15/5 tasks after 4000 training dialogues.}.

Compare FM-DGNN with DQN, we can find that FM-DGNN significantly performs better than DQN in almost all of tasks. 
The set-up of FM-DGNN is same as that of DQN except that the network of FM-DGNN is Dual GNN, while the network of DQN is MLP. The results show that FM-DGNN not only converges much faster but also obtain better final performance. As discussed in section \ref{sec:gnn-policy}, the reason is that with the shared parameters, the skills can be transferred between S-agents, which can improve the speed of learning and the generalization of policy.

Compare FM-DGNN with GP-Sarsa, we can find that the performance of FM-DGNN is comparable to that of GP-Sarsa in simple tasks (e.g. CR-Env.1 and CR-Env.2), while in complex tasks (e.g. LAP-Env.5 and LAP-Env.6) FM-DGNN performs much better than GP-Sarsa. It indicates that FM-DGNN is well suit to large-scale complex domains.

We also compare Feudal-DQN with FM-DGNN in Table \ref{tab:results}\footnote{We find that the performance of Feudal-DQN is sensitive to the order of the actions of master policy in the configuration. Here we show the best results of Feudal-DQN.}. We find that the average performance of FM-DGNN is better than that of Feudal-DQN after both 1000 and 4000 training dialogues. In some tasks (e.g. SFR-Env.1 and LAP-Env.1), the performance of Feudal-DQN is rather low. In \cite{casanueva2018feudal}, authors find that Feudal-DQN is prone to ``overfit'' to an incorrect action. However, here FM-DGNN doesn't suffer from this problem. 

Finally, we compare FM-DGNN with FM-GNN, and find that FM-DGNN consistently outperforms FM-GNN on all tasks. This is due to the \textit{Graph Dueling Operation} (GDO), which implicitly divides a task spatially.

\begin{figure*}
\centering
\includegraphics[width=0.9\textwidth]{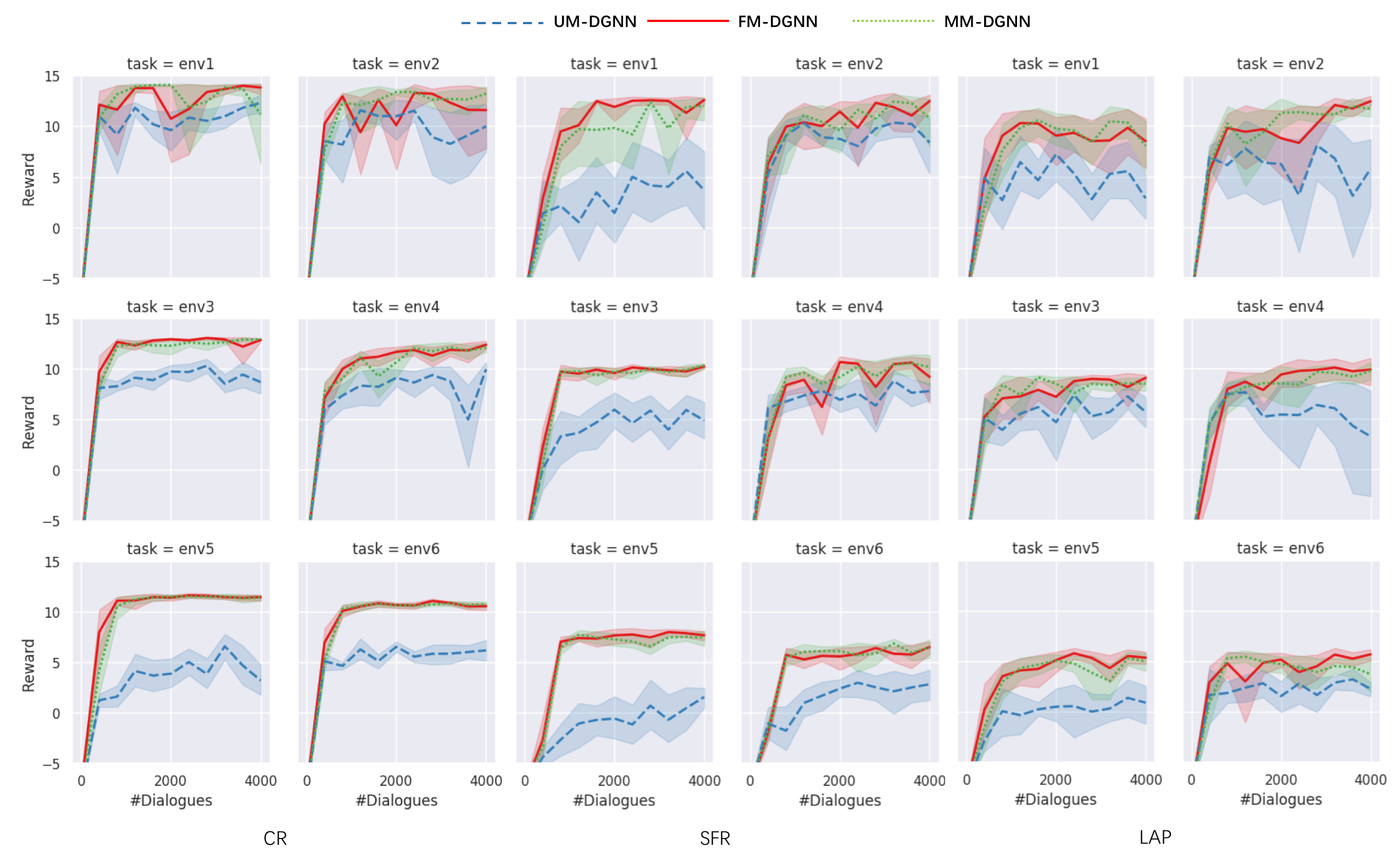}
\caption{The learning curves of reward for the  DGNN-based dialogue policies with three different graph structures, i.e. fully connected (FC) graph, master-node (MN) graph, and isolated (FU) graph. MM-DGNN, UM-DGNN and FM-DGNN are three DGNN-based policies with MN, FU and FU respectively.}
\label{fig:topo-1}
\end{figure*}

\begin{figure*}
\centering
\includegraphics[width=0.9\textwidth]{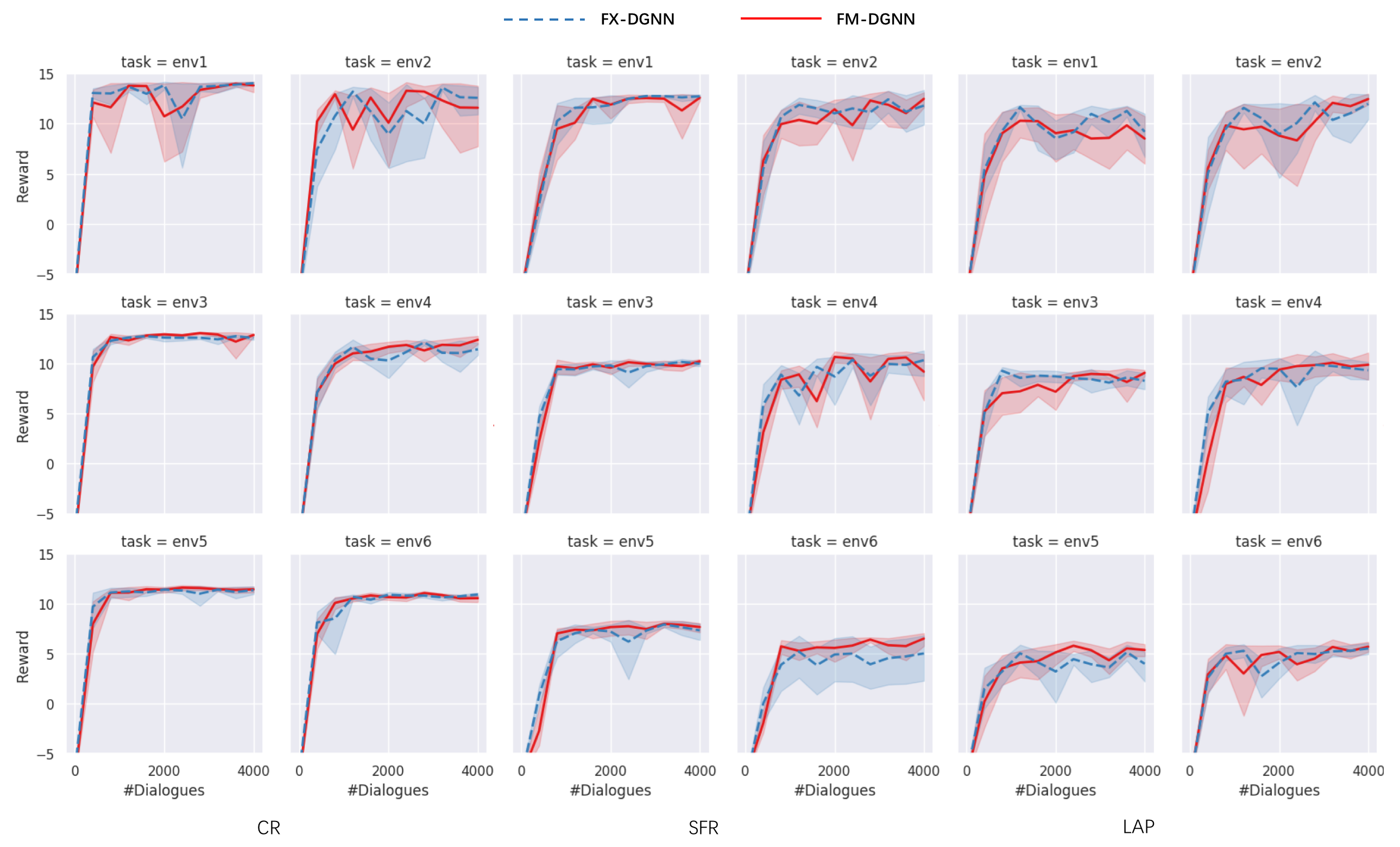}
\caption{The learning curves of reward for the  DGNN-based dialogue policies with two different communication methods, i.e. \textit{Max-Comm} (FX-DGNN) and \textit{Mean-Comm} (FM-DGNN).}
\label{fig:topo-2}
\end{figure*}

\subsection{Effect of Graph Structure  and Communication Method}
\label{sec:exp-structure}

In this subsection, we will investigate the effect of graph structures and communication methods in DGNN-based dialogue policies. 

Fig. \ref{fig:topo-1} shows the learning curves of reward for the DGNN-based dialogue policies with three different graph structures, i.e. fully connected (FC) graph, master-node (MN) graph and isolated (FU) graph. MM-DGNN, UM-DGNN and FM-DGNN are three DGNN-based policies with MN, FU and FC respectively. We can find that FM-DGNN and MM-DGNN perform much better than UM-DGNN on all tasks, which means that message exchange between agents (nodes) is very important. We further compare FM-DGNN with MM-DGNN, and find that there is almost no difference between their performance, which shows that the communication between S-node and I-node is important, while the communication between S-nodes is unnecessary on these tasks. 

In Fig. \ref{fig:topo-2}, we compare the DGNN-based dialogue policy with two different communication methods, i.e. \textit{Max-Comm} (FX-DGNN) and \textit{Mean-Comm} (FM-DGNN). It shows that there is no significant difference between their performance.  This phenomenon has also been observed in other fields, e.g. image recognition \cite{wang2018non}.

\begin{figure*}
\centering
\includegraphics[width=0.9\textwidth]{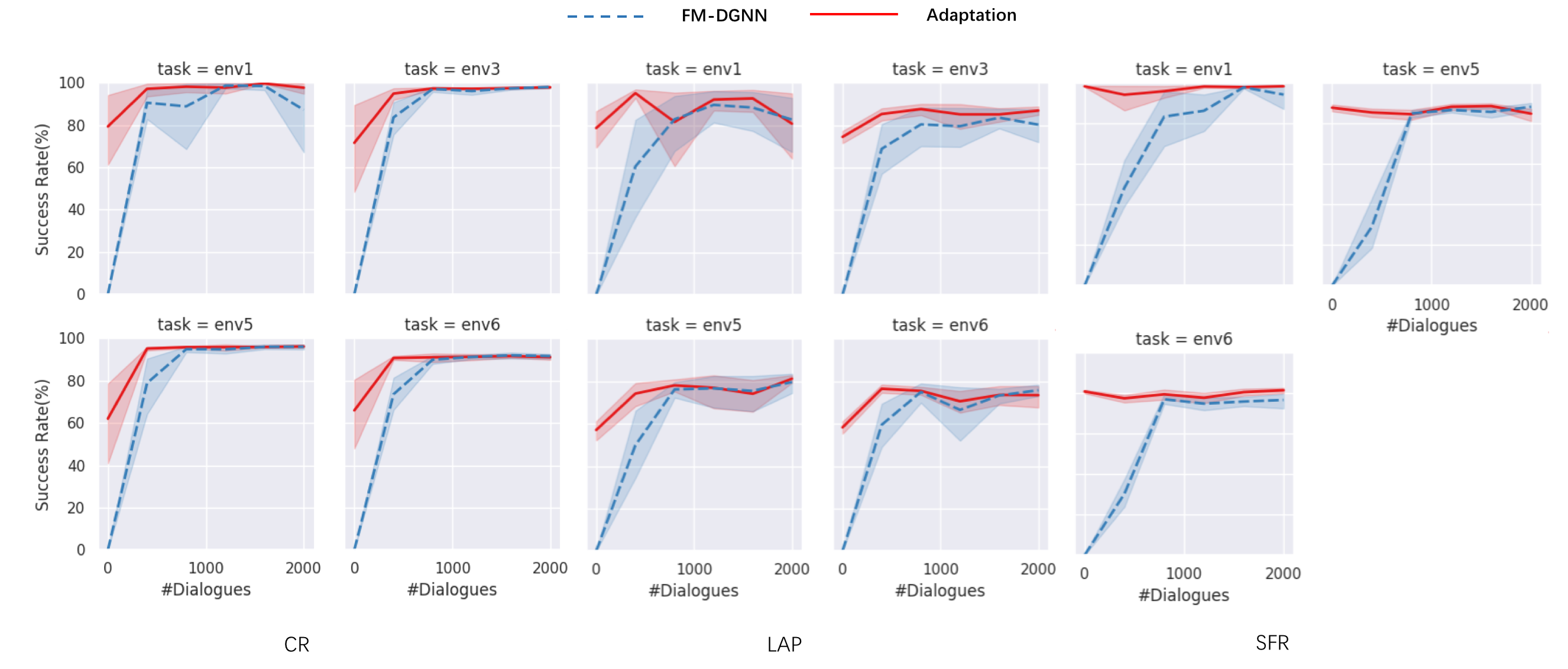}
\caption{The success rate learning curves of FM-DGNN w/o policy adaptation. For adaptation, FM-DGNN is trained on SFR-Env.3 with 4000 dialogues. Then the pre-trained policy is used to initialize the policy on new tasks. For comparison, the policies (orange ones) optimized from scratch are also shown here.}
\label{fig:trans}
\end{figure*}

\subsection{Policy Transfer Learning}
\label{sec:exp-transfer}

\begin{table}
\begin{center}
\caption{Summary of Dialogue Policy Adaptation}
\begin{tabular}{|c|c|c|c|c|}
\hline
 \multirow{2}{*}{Category of Adaptation}
 & \multicolumn{2}{|c|}{Source Task} & \multicolumn{2}{|c|}{Target Task} \\
\cline{2-5}
 & Domain & Env. & Domain & Env. \\
\hline
\multirow{3}{*}{Environment Adaptation} & SFR & Env.3 & SFR  & Env.1 \\
  & SFR & Env.3 & SFR  & Env.6 \\
\cline{2-5} 
 & SFR & Env.3 & SFR  & Env.5 \\
\hline
\multirow{2}{*}{Domain Adaptation}
 & SFR & Env.3 & CR  & Env.3 \\
 & SFR & Env.3 & LAP & Env.3 \\
\hline 
\multirow{6}{*}{Complex Adaptation}
 & SFR & Env.3 & CR & Env.1 \\
 & SFR & Env.3 & CR & Env.5 \\
 & SFR & Env.3 & CR & Env.6 \\
 & SFR & Env.3 & LAP & Env.1 \\
 & SFR & Env.3 & LAP & Env.5 \\
 & SFR & Env.3 & LAP & Env.6 \\
\hline
\end{tabular}
\label{tab:transfer-condition}
\end{center}
\end{table}

In this subsection, we evaluate the adaptability of AgentGraph models. FM-DGNN is first trained with 4000 dialogues on the source task SFR-Env.3, then transferred to the target tasks, i.e on a new task the pre-trained policy is as the initial policy and continue to be trained with another 2000 dialogues. Here we investigate policy adaptation on different conditions, which are summarized in Table \ref{tab:transfer-condition}. The learning curves of success rates on target tasks are shown in Fig. \ref{fig:trans}.  

\subsubsection{Environment Adaptation} 
In real life applications, the conversation agents inevitably interact with new users, the behaviors of which may be different to previous. Therefore, it is very important that the agents have the adaptability to users with different behaviors. In order to test the user adaptability of AgentGraph, we first train FM-DGNN on SFR-Env.3 with standard users, then continue to train the model on SFR-Env.5 with unfriendly users. The learning curve on SFR-Env.5 is shown at the top right of Fig. \ref{fig:trans}. The success rate at 0 dialogues is the performance of the pre-trained policy without fine-tune on the target task. We can find that pre-trained model with standard users performs very well on the task with unfriendly users.

Another challenge in practice for conversation agents is that the input components including ASR and SLU are very likely to make errors. Here we want to evaluate how well AgentGraph can learn the optimal policy in face of noisy input with different semantic error rates (SER). We first train FM-DGNN  under 15\% SER (SFR-Env.3), then continue to train the model under 0\% SER (SFR-Env.1) and 30\% SER (SFR-Env.6) respectively. The learning curves on SFR-Env.1 and SFR-Env.6 are shown in Fig. \ref{fig:trans}. We can find that on both tasks the learning curve is almost a horizontal line. It indicates the AgentGraph model has the adaptability to different level noises.

\subsubsection{Domain Adaptation}
As discussed in section \ref{sec:policy_transfer}, AgentGraph policy can be directly transferred from source domain to another domain, even though the ontologies of two domains are different. Here FM-DGNN is first trained in SFR domain  (SFR-Env.3), then the parameters of FM-DGNN is used to initialize the policies in the CR domain (CR-Env.3) and the LAP domain (LAP-Env.3). It is notable that the task SFR-Env.3 is more difficult than the task CR-Env.3 and simpler than the task LAP-Env.3. The results on CR-Env.3 and LAP-Env.3 are shown in Fig. \ref{fig:trans}. We can find that the initial success rate on both target tasks is more than 75\%. We think this is an efficient way to solve the cold start problem in dialogue policy learning. Moreover, compared with the policy optimized from scratch on the target tasks, the pre-trained policies converge much faster.

\subsubsection{Complex Adaptation}
We further evaluate the adaptability of AgentGraph policy when the environment and the domain both change. Here FM-DGNN is pre-trained with standard users in the SFR domain under 15\% SER (SFR-Env.3), then transferred to other two domains (CR and LAP) in different environments (CR-Env.1, CR-Env.5, CR-Env.6, LAP-Env.1, LAP-Env.5, and LAP-Env.6). The learning curves on these target tasks are shown in Fig. \ref{fig:trans}. We can find that on most of these tasks the policies can obtain acceptable initial performance and converge very fast. They will converge with less than 500 dialogues.

\section{Conclusion}
\label{sec:conclusion}

This paper has described a structured deep reinforcement learning framework, \textit{AgentGraph}, for dialogue policy optimization.  The proposed AgentGraph is the combination of GNN-based architecture and DRL-based algorithm. It can be regarded as one of the multi-agent reinforcement learning approaches. Multi-agent RL has been previously explored for multi-domain dialogue policy optimization \cite{gavsic2015policy}. However, here it is investigated for improving the learning speed and the adaptability of policy in single domains. 
Under AgentGraph framework, we propose a GNN-based dialogue policy and its variant  Dual GNN-based dialogue policy, which implicitly decomposes the decision in each turn into a high-level global decision and a low-level local decision.

Compared with traditional RL approaches, AgentGraph models not only converge faster but also obtain better final performance on most tasks of PyDial benchmark. The gain is larger on complex tasks.
We further investigate the effect of graph structures and communication methods in GNNs. It shows that messages exchange between agents is very important. However, the communication between S-agent and I-agent is more important than that between S-agents. We also test the adaptability of AgentGraph under different transfer conditions. We find that AgentGraph not only has acceptable initial performance but also converges faster on target tasks. 

The proposed AgentGraph framework shows promising perspectives of future improvements. 
\begin{itemize}
\item Recently, several improvements to the DQN algorithm have been made \cite{hessel2018rainbow}, e.g. prioritized experience replay \cite{schaul2015prioritized}, multi-step learning \cite{sutton1988learning}, and noisy exploration \cite{fortunato2017noisy}. The combination of these extensions provides state-of-the-art performance on the Atari 2600 benchmark. 
Integration of these technologies in AgentGraph is one of future work.
\item In this paper, the value-based RL algorithm, i.e. DQN, is used. As discussed in section \ref{sec:sdrl}, in principle other DRL algorithms, e.g. policy-based \cite{sutton2000policy} and actor-critic \cite{mnih2016asynchronous} approaches, can also be used in AgentGraph. We will explore how to combine these algorithms with AgentGraph in our future work. 
\item Our proposed AgentGraph can be regarded as one of \textit{spatial} hierarchical RL, and is used for policy optimization in a single domain. In real-world applications, a conversation may involve multi-domains, which it's challenging to solve for traditional flat RL. Several \textit{temporal} hierarchical RL methods \cite{baolin-hier,budzianowski2017sub}  have been proposed to tackle this problem. Combination of spatial and temporal hierarchical RL methods is an interesting future research direction.

\item In practice, commercial dialog systems usually  involve many business rules, which are represented by some auxiliary variables and their relations. One way to encode these rules in AgentGraph is first to transform rules into \textit{relation graphs}, and then learn representations over them with GNN \cite{allamanis2017learning}.

\end{itemize}

\bibliographystyle{IEEEtran}
\bibliography{paper.bib}

\begin{IEEEbiography}[{\includegraphics[width=1in,height=1.25in,clip,keepaspectratio]{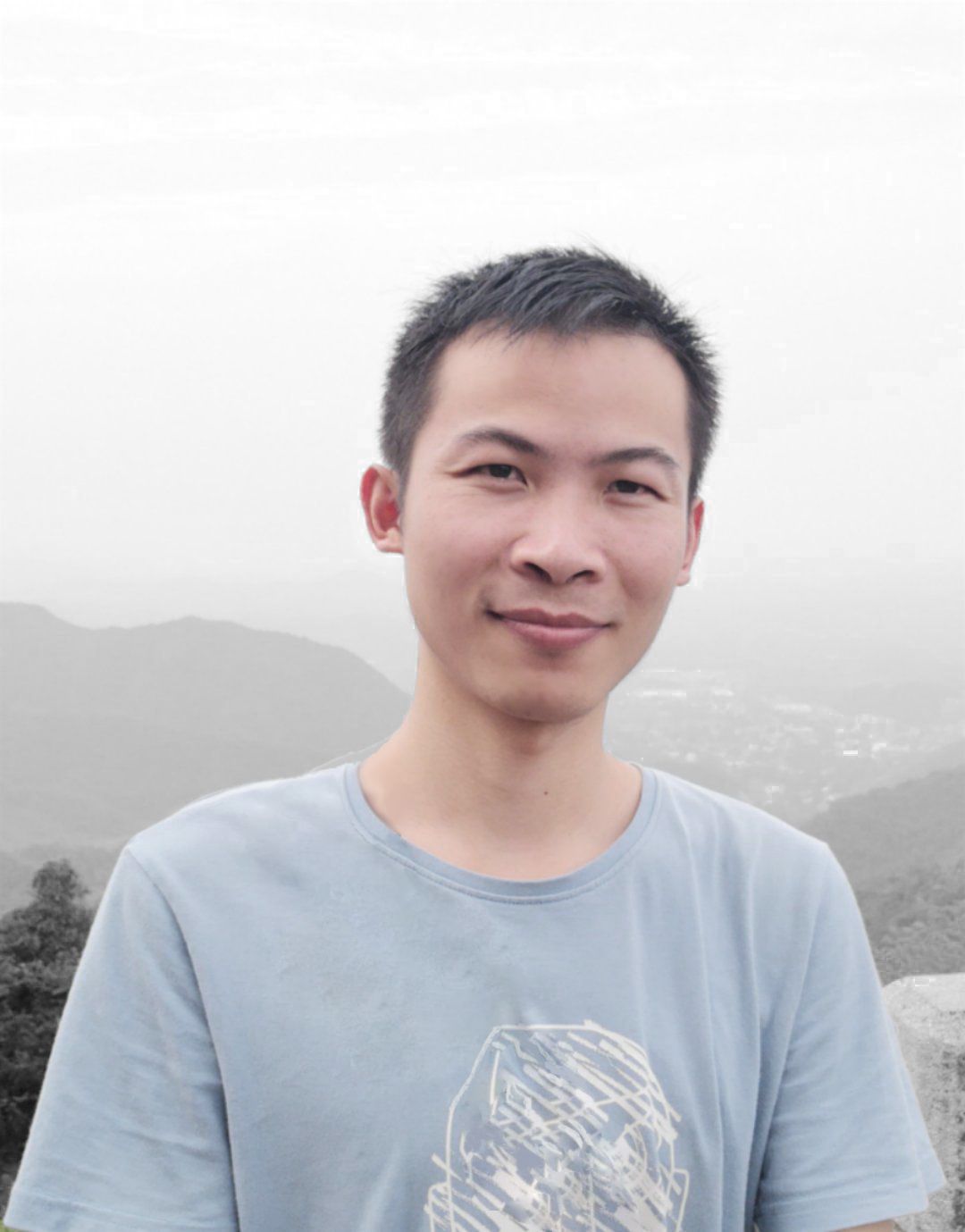}}]{Lu Chen}
received his B.Eng. degree from the School of Computer Science and Technology, Huazhong University of Science \& Technology, in 2013. 
He is currently a Ph.D. student in the SpeechLab, Department of Computer Science and Engineering, Shanghai Jiao Tong University. His research interests include dialogue systems, reinforcement learning, and structured  deep learning. 
\end{IEEEbiography}

\begin{IEEEbiography}[{\includegraphics[width=1in,height=1.25in,clip,keepaspectratio]{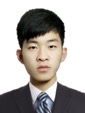}}]{Zhi Chen}
received his B.Eng. degree from the School of Software and Microelectronics, Northwestern Polytechnical University, in 2017. 
He is currently a Ph.D. student in the SpeechLab, Department of Computer Science and Engineering, Shanghai Jiao Tong University. His research interests include dialogue systems, reinforcement learning, and structured  deep learning. 
\end{IEEEbiography}

\begin{IEEEbiography}[{\includegraphics[width=1in,clip,keepaspectratio]{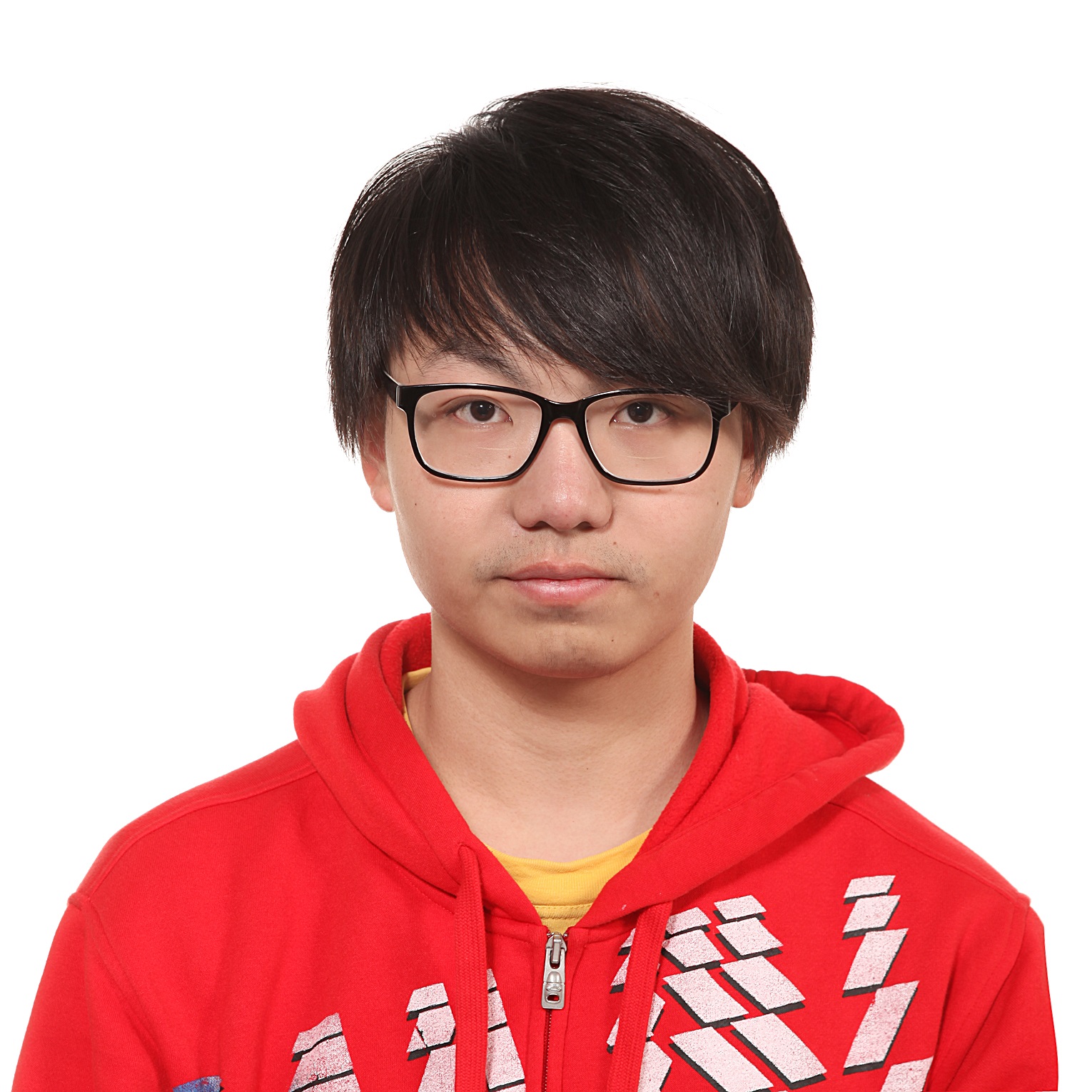}}]{Bowen Tan}
will receive his B.Eng. degree from the Department of Computer Science and Engineering, Shanghai Jiao Tong University, in June 2019. 
He is currently an undergraduate researcher in the SpeechLab, Department of Computer Science and Engineering, Shanghai Jiao Tong University. His research interests include dialogue systems, text generation and reinforcement learning. 
\end{IEEEbiography}

\begin{IEEEbiography}[{\includegraphics[width=1in,clip,keepaspectratio]{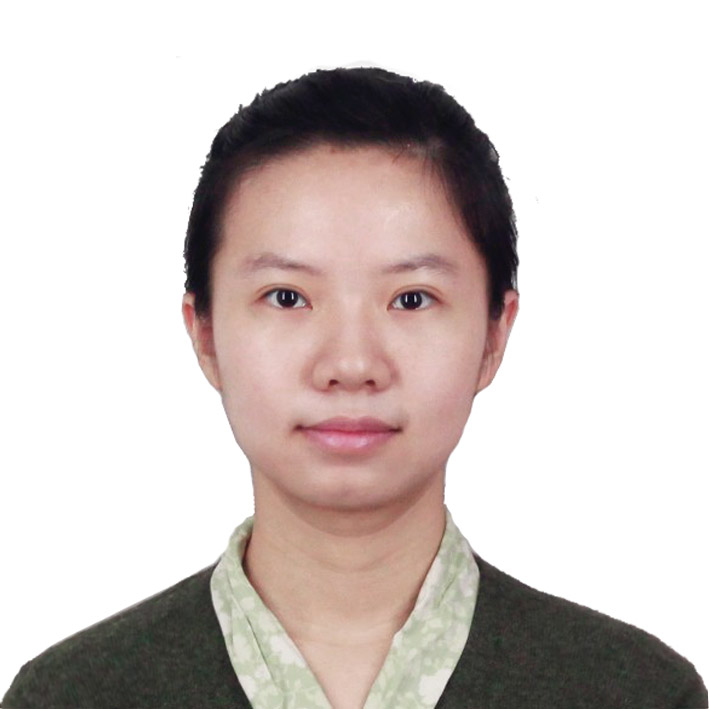}}]{Sishan Long}
will receive her B.Eng. degree from the Department of Computer Science and Engineering, Shanghai Jiao Tong University, in June 2019. 
She is currently an undergraduate researcher in the SpeechLab, Department of Computer Science and Engineering, Shanghai Jiao Tong University. Her research interests include distributed systems and dialogue systems. 
\end{IEEEbiography}

\begin{IEEEbiography}[{\includegraphics[width=1in,height=1.25in,clip,keepaspectratio]{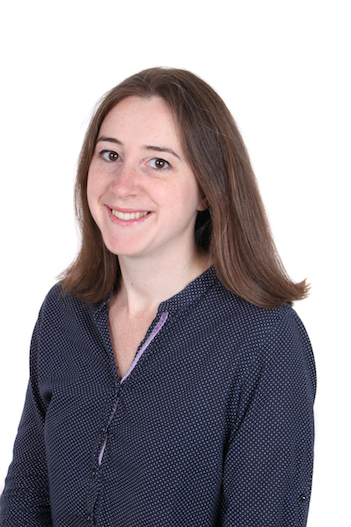}}]{Milica Ga{\v{s}}i{\'c}}
 is a professor at Heinrich-Heine-University D\"{u}sseldorf and holds chair for Dialog Systems and Machine Learning. Prior to that she was a Lecturer at the Cambridge University Engineering Department. She has a BS in Computer Science and Mathematics from the University of Belgrade (2006), an MPhil in Computer Speech, Text and Internet Technology (2007) and a PhD in Engineering from the University of Cambridge (2011). The topic of her PhD was statistical dialogue modelling and she was awarded an EPSRC PhD plus award for her dissertation.  She has published around 60 journal articles and peer reviewed conference papers. 
 Most recently she obtained an ERC Starting Grant and an Alexander von Humboldt award. She is an elected member of IEEE SLTC and appointed board member of Sigdial.
\end{IEEEbiography}

\begin{IEEEbiography}[{\includegraphics[width=1in,height=1.25in,clip,keepaspectratio]{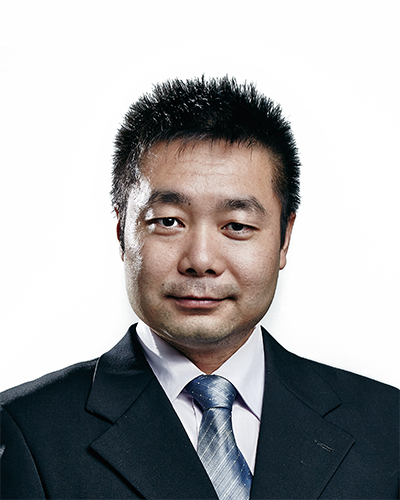}}]{Kai Yu}
 is a professor at Computer Science and Engineering Department, Shanghai Jiao Tong University, China. He received his B.Eng. and M.Sc. from Tsinghua University, China in 1999 and 2002, respectively. He then joined the Machine Intelligence Lab at the Engineering Department at Cambridge University, U.K., where he obtained his Ph.D. degree in 2006. His main research interests lie in the area of speech-based human machine interaction including speech recognition, synthesis, language understanding and dialogue management. He is a member of the IEEE Speech and Language Processing Technical Committee.
\end{IEEEbiography}

\end{document}